%% file: manuscript.tex
\documentclass[a4paper,fleqn]{cas-sc}

\usepackage[authoryear]{natbib}
\usepackage{subcaption}
\usepackage{bm}
\usepackage{amsmath, amsfonts, amssymb}
\usepackage{multirow}
\usepackage[outline]{contour}

\usepackage{booktabs,tabularx, longtable}
\usepackage{placeins}

\usepackage{xcolor}

\usepackage[normalem]{ulem} 

\usepackage[markup=underlined,
            authormarkup=brackets,
            commandnameprefix=ifneeded]{changes}

\definechangesauthor[name=Hongrui, color=teal]{HS}

\def\tsc#1{\csdef{#1}{\textsc{\lowercase{#1}}\xspace}}
\tsc{WGM}
\tsc{QE}

\usepackage[left,mathlines]{lineno}

\begin{document}


\let\WriteBookmarks\relax
\def\floatpagepagefraction{1}
\def\textpagefraction{.001}

\shorttitle{}    

\shortauthors{}  

\title [mode = title]{Habitat Classification from Ground-Level Imagery Using Deep Neural Networks}  





%

\author[1]{Hongrui Shi}[orcid=0000-0002-3075-2639,]

\cormark[1]


\ead{hshi@lincoln.ac.uk}


\credit{Conceptualization, Methodology, Software, Validation, Formal analysis, Investigation, Data curation, Writing - Original Draft, Writing - Review \& Editing, Visualization}
\author[2]{Lisa Norton}
\credit{Conceptualization, Validation, Resources, Data curation, Writing - Review \& Editing, Supervision, Project administration, Funding acquisition}
\author[2]{Lucy Ridding}
\credit{Validation, Resources, Data curation, Writing - Review \& Editing, Project administration}
\author[3]{Simon Rolph}
\credit{Resources, Writing - Review \& Editing, Project administration}
\author[3]{Tom August}
\credit{Resources,  Writing - Review \& Editing, Funding acquisition}
\author[2]{Claire M Wood}
\credit{Resources,  Data curation}
\author[4]{Lan Qie}
\credit{Conceptualization, Validation,  Writing - Review \& Editing, Supervision, Project administration, Funding acquisition}
\author[1, 5]{Petra Bosilj}
\credit{Conceptualization, Methodology, Resources,  Writing - Review \& Editing, Supervision, Project administration, Funding acquisition}
\author[1]{James M Brown}
\credit{Conceptualization, Methodology, Resources,  Writing - Review \& Editing, Supervision, Project administration, Funding acquisition}

\affiliation[1]{organization={School of Engineering \& Physical Sciences, University of Lincoln},
            addressline={Brayford Pool}, 
            city={Lincoln},
            postcode={LN6 7TS}, 
            country={United Kingdom}}
            
\affiliation[2]{organization={UK Centre for Ecology \& Hydrology},
            addressline={Lancaster Environment Centre, Library Avenue, Bailrigg}, 
            city={Lancaster},
            postcode={LA1 4AP}, 
            country={United Kingdom}}
            
\affiliation[3]{organization={UK Centre for Ecology \& Hydrology},
            addressline={Maclean Building, Benson Lane, Crowmarsh Gifford, Wallingford},
            city={Oxfordshire},
            postcode={OX10 8BB}, 
            country={United Kingdom}}
            
\affiliation[4]{organization={School of Natural Sciences, University of Lincoln},
            addressline={Brayford Pool}, 
            city={Lincoln},
            postcode={LN6 7TS}, 
            country={United Kingdom}}

\affiliation[5]{organization={Department of Advanced Computing Sciences, Maastricht University},
            city={Maastricht},
            postcode={6229 GS}, 
            country={The Netherlands}}

\cortext[1]{Corresponding author}



\begin{abstract}
Habitat assessment at local scales --- critical for enhancing biodiversity and guiding conservation priorities --- often relies on expert field surveys that can be costly, motivating the exploration of AI-driven tools to automate and refine this process. While most AI-driven habitat mapping depends on remote sensing, it is often constrained by sensor availability, weather, and coarse resolution. In contrast, ground-level imagery captures essential structural and compositional cues invisible from above and remains underexplored for robust, fine-grained habitat classification. This study addresses this gap by applying state-of-the-art deep neural network architectures to ground-level habitat imagery. Leveraging data from the UK Countryside Survey covering 18 broad habitat types, we evaluate two families of models --- convolutional neural networks (CNNs) and vision transformers (ViTs) --- under both supervised and supervised contrastive learning paradigms. Our results demonstrate that ViTs consistently outperform state-of-the-art CNN baselines on key classification metrics (Top-3 accuracy = 91\%, MCC = 0.66) and offer more interpretable scene understanding tailored to ground-level images. Moreover, supervised contrastive learning significantly reduces misclassification rates among visually similar habitats (e.g., Improved vs. Neutral Grassland), driven by a more discriminative embedding space. Finally, our best model performs on par with experienced ecological experts in habitat classification from images, underscoring the promise of expert-level automated assessment. By integrating advanced AI with ecological expertise, this research establishes a scalable, cost-effective framework for ground-level habitat monitoring to accelerate biodiversity conservation and inform land-use decisions at a national scale. 
\end{abstract}



\begin{keywords}
habitat classification \sep environmental monitoring \sep deep learning \sep vision transformers \sep conservation \sep biodiversity \sep ground-level imagery
\end{keywords}

\maketitle











\input{introduction}
\input{method}
\input{experiments}

\input{conclusion}

\printcredits

\section*{Funding}
This work was funded by Natural Environment Research Council of UK Research and Innovation [grant numbers UKRI054]. 

\section*{Declaration of Generative AI and AI‐assisted Technologies in the Writing Process}
During the preparation of this work the author(s) used ChatGPT in order to improve language and readability. After using this tool/service, the author(s) reviewed and edited the content as needed and take(s) full responsibility for the content of the publication.

\section*{Data Availability}
The code used for model training and evaluation is publicly available at \url{https://github.com/WhiteGiveFive/Habitat-Classification-from-Ground-Level-Imagery} (MIT license). The data used in this study are not publicly hosted due to UKCEH's data privacy policy but may be obtained from UKCEH under a research licence upon request. Requests for access to the ground-level photographs and corresponding UKHab labels should be directed to Dr Lisa Norton (\texttt{lrn@ceh.ac.uk}).





\bibliographystyle{cas-model2-names}

\bibliography{cas-refs}

\clearpage
\input{appendix}

\end{document}

%% file: introduction.tex
\section{Introduction}




Habitats are generally classified according to underlying biophysical characteristics of the land on which they occur and the associated species assemblages. Accurate habitat classification is essential for biodiversity conservation and effective land management. Habitat extent and condition data underpin policy frameworks—such as UK's Biodiversity Net Gain \citep{zu2021exploring, defra2023bng} and EU Biodiversity Strategy for 2030 \citep{ec2020biodiv2030}—and guide conservation priorities, yet they currently depend on expert field surveys which can be costly or difficult to access, for example for landowners looking to understand their land and its value in relation to offsetting or carbon credits \citep{schofield2024aren}. Recent advances in artificial intelligence (AI) have demonstrated strong potential to automate image-based ecological classification tasks, making habitat identification more accessible for non-experts, thereby reducing costs and alleviating professional workloads. 


To date, most AI-driven habitat mapping works rely on remote sensing data (e.g., satellite or drone imagery) \citep{van2023high, norris2024comparing}. Though remote-sensing-based habitat classification excels at quantifying spatial extent, it can be hindered by sensor availability, weather conditions, and relatively coarse spatial resolution (satellite imagery typically several m/pixel) \citep{diaz2024classification, morueta2024unlocking, marjani2025novel}. Ground-level imagery, by contrast, captures local structural and compositional cues for habitats \citep{perrett2023deepverge}, which are invisible from above but critical for precise habitat classification. It is effective for site-scale habitat assessment and species identification and can be collected with consumer devices rather than a complex remote-sensing pipeline. However, this approach has been less studied for AI-driven habitat classification. In the UK, image analysis is widely used to help both expert and non-expert field surveyors identify species, but no such equivalent currently exists for habitats. Yet, habitat classification is important for mapping and decision-making regarding land use and management, as well as in the context of BNG, a government supported approach currently in use in England to ensure that new projects deliver measurable improvement in nature either on site, or in areas away from the development site (offsets). \citep{mall2023remote, morueta2024unlocking} In this context, we approach automatic habitat classification using ground-level imagery. By leveraging state-of-the-art AI techniques on richly annotated images from the UK Countryside Survey \citep{carey2008countryside}, we seek to support cost-effective site-level habitat monitoring, land cover mapping, and complement established approaches such as remote sensing.



Recently, data-driven computer vision has become instrumental in ecological and environmental research by leveraging the growing volume of imagery from diverse sensors to deliver social and economic benefits \citep{reynolds2025potential, oba2025accelerating}. Applications include species detection and monitoring, where deep learning models automate identification and population assessment from camera traps \citep{ratnayake2021tracking, bjerge2022real, xu2024advancing}; underwater and benthic habitat surveys \citep{zhong2023fine}, which leverage images recorded by underwater cameras to monitor coral reef growth; plant recognition \citep{joly2024overview, wang2025plant}, employing large public datasets to train deep neural networks to classify plant species (e.g. PlantCEF \citep{plantnet2025}); and habitat classification \citep{van2023high, norris2024comparing}.

Within these applications, habitat classification is most directly aligned with our work. We therefore review it in more detail, focusing on data sources and model architectures where research gaps emerge. Currently, habitat classification remains particularly reliant on remote sensing, including satellite data~\citep{pratico2021machine, sittaro2022machine, marcinkowska2023natura, van2023high} and drone-based images~\citep{retallack2022using, norris2024comparing}. While these sources enable coverage of large areas, they depend on favourable environmental conditions, require specialised devices and expertise, and focus on a single top-down perspective, which limits their practicality for non-experts seeking timely, user-friendly, and precise habitat evaluations~\citep{morueta2024unlocking}. Compared to satellite imagery, ground-level imagery can be easily captured by on-site surveyors or even citizen scientists using consumer devices, offering a cost-effective alternative for detailed habitat classification. For example, DeepVerge~\citep{perrett2023deepverge} demonstrates that street-view images can be used to train CNNs for counting grassland species. \cite{saadeldin2022using} shows that a simple CNN trained on a limited number of ground-level imagery achieves accurate classifications over three categories of grassland management intensity. \cite{martinez2024automatic} applies semantic segmentation to decompose visual features in ground-level photos and uses these features to classify coarse habitat classes (e.g., grassland, cropland), equivalent to UKHab Level 2 categories. However, despite this potential, ground-level imagery remains underexplored to address habitat classification across a variety of subcategories (e.g. different grasslands, woodlands, wetlands).

Regarding the adopted models for habitat classification, previous studies~\citep{gomez2019towards, pratico2021machine, sittaro2022machine, marcinkowska2023natura, van2023high} have predominantly used classical machine learning models, such as support vector machines~\citep{cortes1995support} and random forests~\citep{breiman2001random}, as well as deep neural networks, specifically CNNs, such as VGG \citep{simonyan2014very} and ResNet \citep{he2016deep}. While these models are effective, they can struggle to address the increasing challenges of learning from more complex images and the growing demand for generalising to a wide range of habitat types. Conversely, the Vision transformer (ViT)~\citep{dosovitskiy2020image} has emerged as a state-of-the-art alternative to CNNs in various computer vision tasks. Compared to the CNN that only extracts isolated regional features, the self-attention mechanism~\citep{vaswani2017attention} --- the foundation of ViT --- allows this model to capture relationships between visual features across the entire image, leading to superior performance and interpretability in many applications. Recent habitat-focused studies incorporating ViTs~\citep{jamali2023wetmapformer, diaz2024classification, marjani2025novel} have shown promising results, but remain limited to a narrow set of habitats (e.g., grassland, wetland) and typically use satellite imagery rather than ground-level photographs. Consequently, there is a clear opportunity to adapt the ViT to more comprehensive, ground-level habitat classification. 

This work aims to develop a model that is capable of automatically classifying diverse habitats based on ground-level imagery and with the long-term vision of deploying it on end-user devices in real-world settings (deployment itself falls beyond the scope of this paper). We leverage deep learning approaches, which have become dominant in various computer vision applications, to achieve this goal. A deep neural network is trained on thousands of labelled ground-level images provided by UK Centre for Ecology \& Hydrology (UKCEH) Countryside Survey (CS) \citep{carey2008countryside}, which represents a wide variety of environments across the UK. Based on the UK Habitat (UKHab) Classification \citep{UKHab2023} with a hierarchical habitat system, each image is annotated by experts with one of the level 3 (L3) habitat subcategories--fine-grained subdivisions of level 2 (L2) habitats (for example, grassland at L2 is split into types such as Acid Grassland, Improved Grassland, etc.). Appendix C provides detailed descriptions of L2 and L3 habitats as defined in the UKHab.



Two challenges relevant to ground-level habitat images are identified in the model development: 1) The model needs to classify habitats based on broad, open scenes with the presence of potentially distracting image features. Ground-level habitat images typically display large, continuous regions (e.g., swaths of grassland) that require an expansive, context-aware understanding of the scene. 2) Some habitat types (for example, Neutral Grassland and Improved Grassland) are visually similar, sharing subtle visual characteristics that makes them difficult to discern. Our training approach must learn distinct feature representations for such habitats so the classifier can differentiate them accurately. Figure \ref{fig: intro, all habitats} displays habitat samples in the CS dataset, as well as the visual challenges of developing the model for automatic habitat classification. 

\begin{figure*}
  \begin{center}
    \includegraphics[width=1.\linewidth]{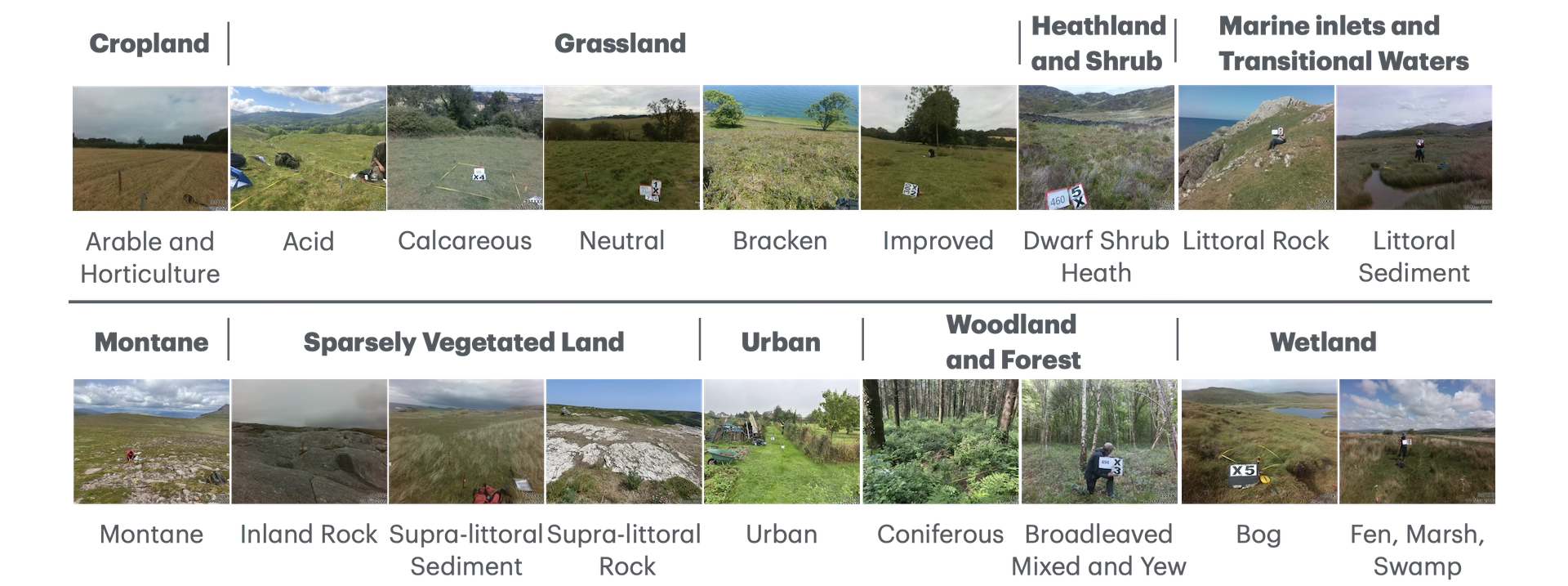}
  \end{center}
  \caption{Examples of level 3 (L3) habitats defined by UKHab in the Countryside Survey dataset, grouped by their coarse L2 categories (bold text). Some L2 habitats, such as Cropland, only have one L3 class. Note that Improved Grassland, Montane, and Bracken have different origins than L3 habitats in UKHab but are treated as L3 habitats in the CS dataset. An explanation is provided in Section \ref{sec: method data}.}
  \label{fig: intro, all habitats}
\end{figure*}

To explore how best to meet these challenges, we structured our methodology around two axes of exploration: 

\begin{itemize}
    \item \textbf{Model architectures.} Convolutional Neural Networks (CNNs) are adopted as our baseline, reflecting the prevailing deep learning model in image classification. We further study Vision Transformer (ViT)~\citep{dosovitskiy2020image}, whose self-attention mechanism~\citep{vaswani2017attention} can more naturally capture global context in ground-level photographs. 
    \item \textbf{Training paradigms.} Supervised learning with a standard cross-entropy objective is first employed for the multiclass habitat classification problem, to establish performance baselines for each architecture. We then investigate Supervised Contrastive Learning (SupCon)~\citep{khosla2020supervised} to pretrain the image encoder, aiming to learn tighter, more separable representations for visually similar habitats.
\end{itemize}

In summary, our research makes four main contributions:
\begin{itemize}
    \item Ground-level habitat classification benchmark. Our benchmark leverages ground-level imagery to address a gap in tools requiring habitat classification (including carbon footprinting, biodiversity offsetting, etc.) for non-experts. It expands the ecological impact of AI by providing more accessible habitat assessments, allowing for extensive habitat classification across the country.
    \item Model architecture finding. Our systematic comparison of representative CNNs and ViTs demonstrates that ViTs consistently outperform comparable CNNs under a broad range of evaluation metrics for habitat classification.
    \item Training paradigm investigation. We investigate a two-stage supervised contrastive (SupCon) learning paradigm (contrastive pretraining and classifier fine-tuning), which delivers modest gains over classical supervised learning for our best-performing model and reduces major confusions (e.g., Improved and Neutral Grassland), thereby improving reliability for real-world deployment.
    \item Human expert comparison. We compare our best model with three experienced ecologists. On a stratified subset, ViT combining with SupCon is competitive with the best human expert in interpreting habitats from level 3 habitat photographs, while revealing limits and opportunities for our approach.

\end{itemize}

%% file: method.tex
\section{Materials and Methods}
\subsection{Data} \label{sec: method data}
Our study focuses on the habitats of the United Kingdom. The habitat images come from the UK Countryside Survey (CS) which covers a randomly stratified sample of UK countryside \citep{wood2017long}, meaning that habitats are sampled in proportion to their occurrence in the wider countryside. Images used were from the most recent rolling survey (CS2019-2023), which covered 500 1km squares across 5 years, during which between 10 and 16 individuals (in each year) were employed as field surveyors. Around half of field surveyors are permanent staff members at UKCEH with other expert surveyors brought in as required. Field surveyors record species within vegetation plots between May and September of the survey year and take photographs of the plots within the context of the habitat in which they are placed, primarily for plot relocation purposes. Specifically, the surveyor is advised to record the location of the plot by taking several ground-level photographs at a near-earth height (1–2 m above ground) using smartphones, including plot markers in the foreground, the wider context of the plot and any significant (easily refindable and preferably permanent features) in the background where possible. These photographs depict the habitat context in the plot. Plots are assigned to habitats by the field surveyors according to the Broad and Priority Habitat classifications \citep{jncc_bap_descriptions}. Broad Habitats comprise part of the hierarchical category system described in the UK Habitat Classification (UKHab) \citep{UKHab2023}. 

UKHab defines five habitat levels, numbered 1 to 5, with higher levels (e.g., Level 3) representing more fine-grained habitat classifications than lower ones (e.g., Level 2). In this study, we particularly focus on classification at level 3 (L3) as level 1 (L1) and level 2 (L2) classifications are too coarse to add practical values for automatic habitat classification and level (L4) was considered too fine-grained. However, L2 classes are still used in this work to describe grouped performance on L3. UKHab is one of several habitat classifications used in the UK but was chosen because it provides a comprehensive unified approach for all UK countries and has been developed to ensure consistency in classification over time and space which is suitable for digital mapping. 

Initially, the data was cleaned by removing unlabelled habitat images, resulting in a dataset with a total number of 5598 samples, including 17 Broad L3 classes and one L4 class (Bracken), corresponding to 9 coarse L2 classes, as shown in Figure \ref{fig: intro, all habitats}. While Bracken is technically defined as an L4 habitat in UKHab, it is also classified as a Broad Habitat and is labelled by field surveyors in the CS dataset alongside the L3 habitats. Improved Grassland and Montane are originally Broad Habitats defined in the UK BAP Priority Habitats system \citep{jncc_bap_habitats}, but they correspond to the scope of L3 habitats in UKHab and are utilised by field surveyors in the CS dataset. For simplicity and consistency throughout this study, we treat Bracken, Improved Grassland, and Montane as equivalent to L3 classes in UKHab, and all references to 'L3 habitats' in the CS dataset include these three classes unless otherwise specified. 

\begin{figure*}
  \centering
    \begin{subfigure}{.48\linewidth}
    \centering
    \centerline{\includegraphics[width=\linewidth]{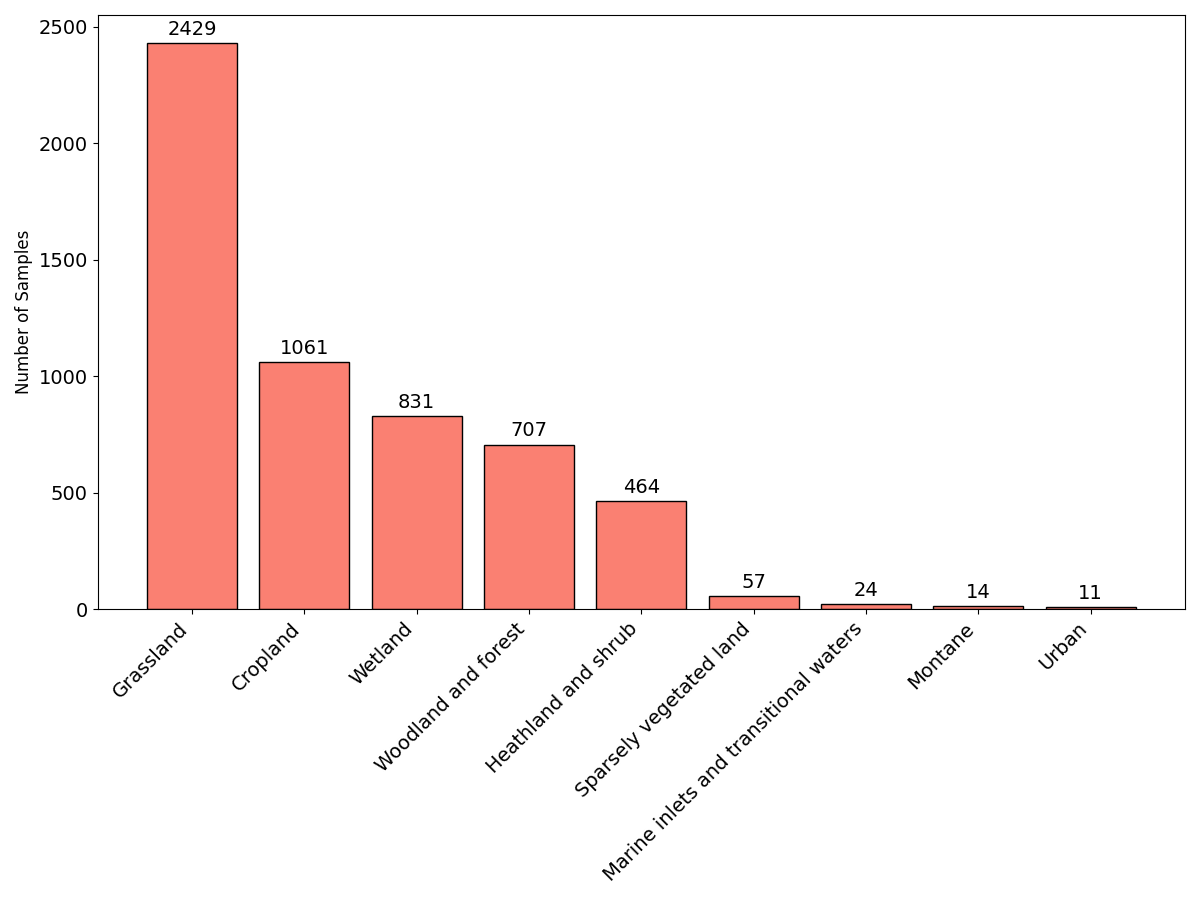}}
    \caption{L2 habitat distribution}
    \end{subfigure}
    \begin{subfigure}{.48\linewidth}
    \centering
    \centerline{\includegraphics[width=\linewidth]{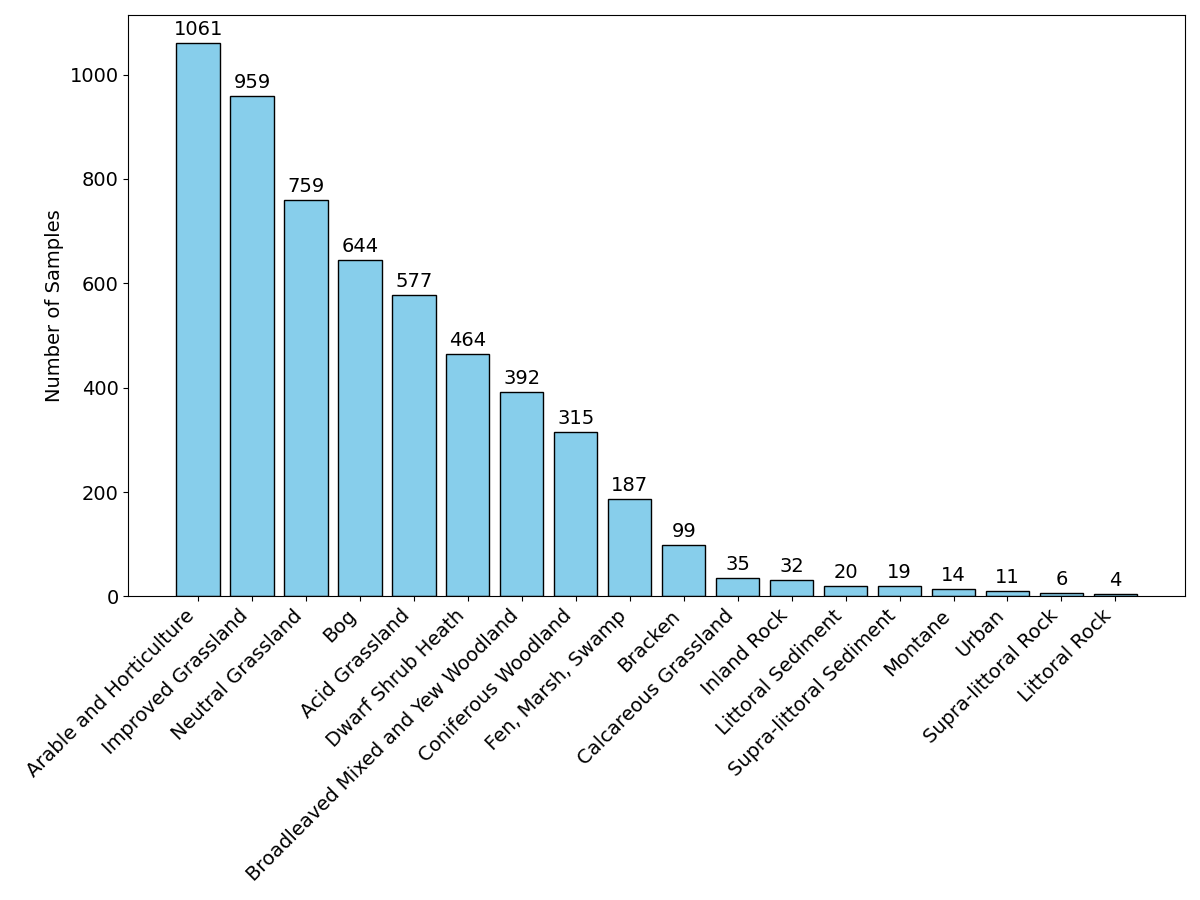}}
    \caption{L3 habitat distribution}
    \end{subfigure} 
  \caption{Habitat distributions (L2 and L3) in the CS dataset based on the UKHab system.}
  \label{fig: CS dataset distributions}
\end{figure*}

Figure~\ref{fig: CS dataset distributions} further depicts the distributions of the habitats at L3 and L2 levels. It shows that the CS dataset has a highly-tailed distribution. The grassland, cropland, wetland, woodland, and heathland are the most common habitats, constituting 98\% of the dataset. We will refer to these habitats as \textit{major habitats}. 


\subsection{Models} \label{sec: models}
The backbone of our habitat classification system is a deep neural network capable of extracting discriminative features from ground-based imagery. In this work, we compare two architectures that have achieved state-of-the-art performance in image classification: conventional convolutional neural networks (CNNs) and Vision Transformers (ViTs). On our CS dataset, each model is trained to map raw image input to habitat labels by learning representations that capture visual patterns in diverse environments. 

\subsubsection{Model Architecture}

\paragraph{Convolutional neural networks.} The convolutional operation is the core of a CNN for modelling images. In particular, convolutional filters represented by matrices slide across the input image and perform element-wise multiplication and summation to detect visual features, such as edges, shapes, and colours \citep{goodfellow2016deep}. The convolutional operation processes the images in local patches. Each time it is calculated, only a small region on the image is activated, whereas the remaining areas are not taken into account. This locality bias assumes that spatially proximate pixels are highly correlated, enabling efficient extraction of meaningful visual features. However, in ground-level habitat imagery---where scenes often contain broad scenes and multiple, potentially distracting objects---this local focus may limit the ability of a CNN to match interpretable visual features to the correct label. 

\paragraph{Vision Transformers.} In contrast, ViT is capable of capturing long-range contextual relationships in the image with the attention mechanism. Specifically, ViT decomposes the input image into a sequence of patches. Each patch represents a small region of the input image and is encoded by the model into a compact representation. To model the relationship of the patches, the attention computes pairwise interactions among all patch representations as follows: 

\begin{equation}
    \text{Attention}(\mathbf{Q}, \mathbf{K}, \mathbf{V}) = \operatorname{softmax}\left(\frac{\mathbf{Q}\mathbf{K}^\top}{\sqrt{d}}\right) \mathbf{V}
    \label{eq: self-attention}
\end{equation}

Here, \(\mathbf{Q}\), \(\mathbf{K}\), and \(\mathbf{V}\) are the query, key matrix, and value matrices obtained from linear projections of the patch representations. The matrix product $\mathbf{Q}\mathbf{K}^\top$ essentially calculates pairwise inner products among projected representations, measuring their similarity. A higher similarity between two patch representations indicates they have a stronger connection. In this way, a broad scene with uniform visual features---such as grasses, woods, or crops---across the entire image can be connected together to represent the habitat. We hypothesise that the global perception of ViT is particularly advantageous for our habitat classification task: compared to the locality bias of CNNs, the non-local interaction enables ViT to learn image representation that encodes long-range regional dependencies and the overall contextual information of the scene, which are critical for generalising on ground-level imagery and interpreting the model performance. 



Figure~\ref{fig: model rationale, cnn vs vit} illustrates the differences of CNN and ViT on modelling images using an exemplary woodland image from the CS dataset.

\begin{figure*}
  \begin{center}
    \includegraphics[width=0.9\linewidth]{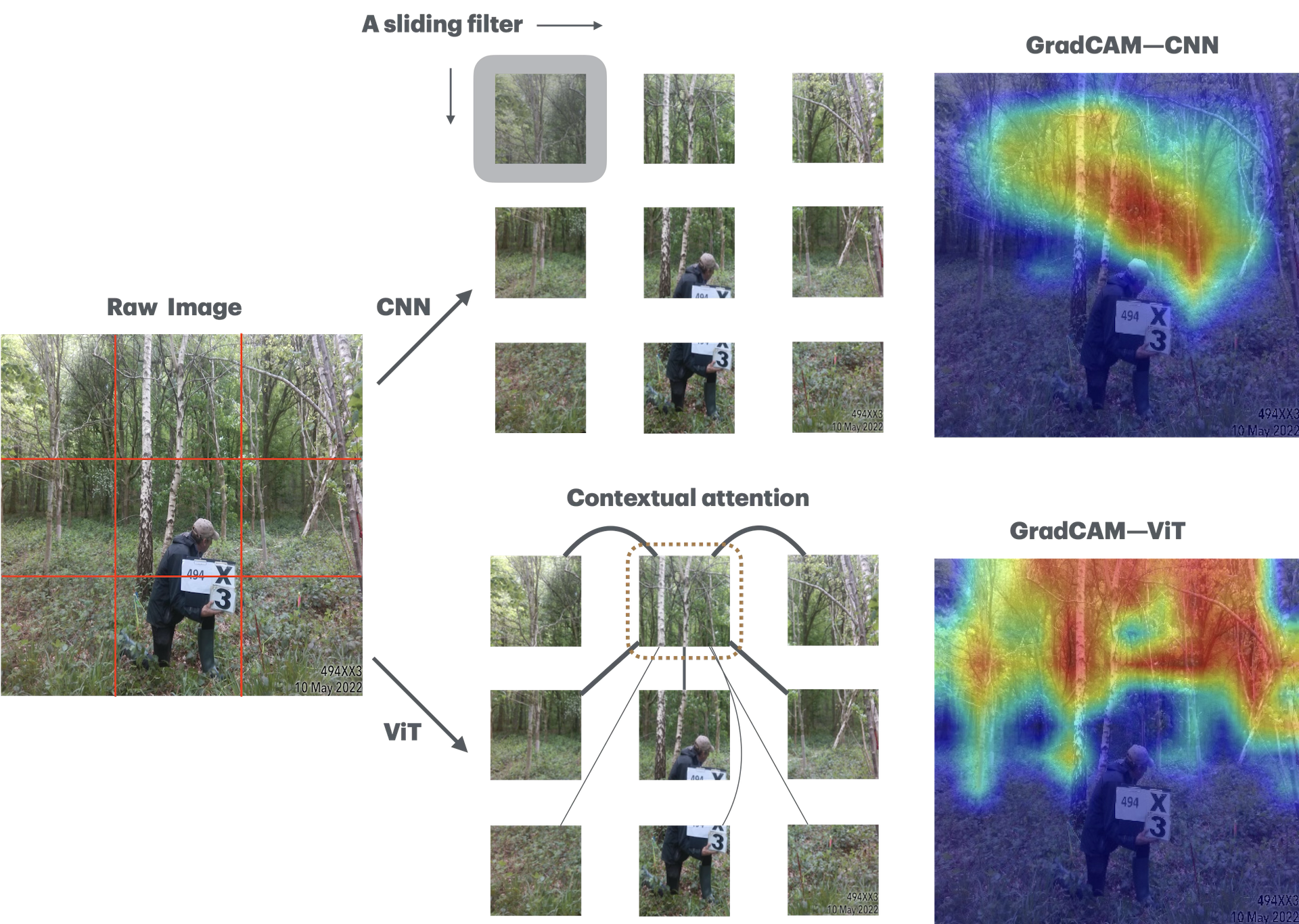}
  \end{center}
  \caption{The conceptual difference between how CNN and ViT process the same ground-level habitat image. A ground-level photograph of a mixed woodland scene is partitioned into a conceptual 3×3 grid of patches. In the top row, the CNN extracts visual features using a sliding filter that only sees one patch at a time, making it tend to focus on prominent features in local regions. This locality is reflected in the GradCAM heatmap, where the model concentrates on the trees in the central region (heated area). In contrast, the ViT (bottom row) allows each patch to “attend” to every other patch through self-attention links. The lines in the diagram are a schematic way to show this idea—the thicker lines simply indicate stronger relationships between patches. The ViT therefore considers the global context of the scene, linking near and far regions. The GradCAM heatmap of ViT highlights areas not only on the central trees but also those along the edges, reflecting a broader contextual understanding. This behaviour aligns more closely with human interpretation of ground-level habitat photos, where habitat context often spans the entire scene rather than a single region.} 
  \label{fig: model rationale, cnn vs vit}
\end{figure*}

\subsubsection{Studied Models} 

We selected representative CNN and ViT models that balance performance and efficiency for the architecture comparison. Using a size sweep to cover a range of capacities, we compared 9 models in total. With respect to the ViT architecture, we employed the Swin Transformer (SwinT)~\citep{liu2021swin} for its computational efficiency. Compared to the basic ViT~\citep{dosovitskiy2020image}, SwinT proposes the shifted window scheme to significantly reduce the computational workload of self-attention, while achieving classification performance on par with or superior to basic ViTs. SwinT has become a standard representative ViT baseline for benchmarking in a number of state-of-the-art computer vision studies \citep{liu2022convnet, zhu2023biformer, wang2023internimage}. The efficiency and capability of SwinT enable us to accelerate the prototyping. 

For the CNN baselines, we utilised three strong CNN models with complementary design principles, each of which has consistently ranked among the top performers on public image classification benchmarks such as CIFAR~\citep{krizhevsky2009learning} and ImageNet~\citep{deng2009imagenet}. They are 1): Wide Residual Network (WRN)~\citep{zagoruyko2016wide}, which increases the number of filters to improve the performance of the popular ResNet~\citep{he2016deep}; 2) EfficientNetV2~\citep{tan2021efficientnetv2}, developed by neural architecture search (NAS) and offering better speed–accuracy trade-offs on ImageNet than both larger CNNs and the basic ViT; and 3) ResNeXt~\citep{xie2017aggregated} introduces cardinality—the number of parallel paths in each block—as a new dimension for scaling CNNs, and achieves superior performance to ResNet on image classification benchmarks.

Additionally, we varied the sizes of the chosen models to provide a more comprehensive comparison, as well as guide our future model deployment on user-end devices, where lightweight models with lower resource demands and competitive capabilities will be favoured. Specifically, SwinT includes size-based variants of small, base, and large models, denoted as SwinT-S, SwinT-B, SwinT-L. In terms of CNNs, WRN-50-2, WRN-101-2, EfficientNetV2-M, EfficientNetV2-L, ResNeXt-50, ResNeXt-101 are chosen for the CNN baselines. Our choices of the ViT and CNN models are not exhaustive, but they are all proven models and can serve well as representatives for their family, enabling robust studies for our habitat classification task.


We followed the widely-accepted pretraining practice to avoid training the model from scratch. All of the chosen models are pretrained on the ImageNet dataset~\citep{deng2009imagenet} and open-weighted. Our pretrained CNNs and SwinT are downloaded from the TorchVision library and the model repository of Hugging Face respectively. 


\subsection{Training Methods} \label{sec: training methods}

\begin{figure*}
  \begin{center}
    \includegraphics[width=0.95\linewidth]{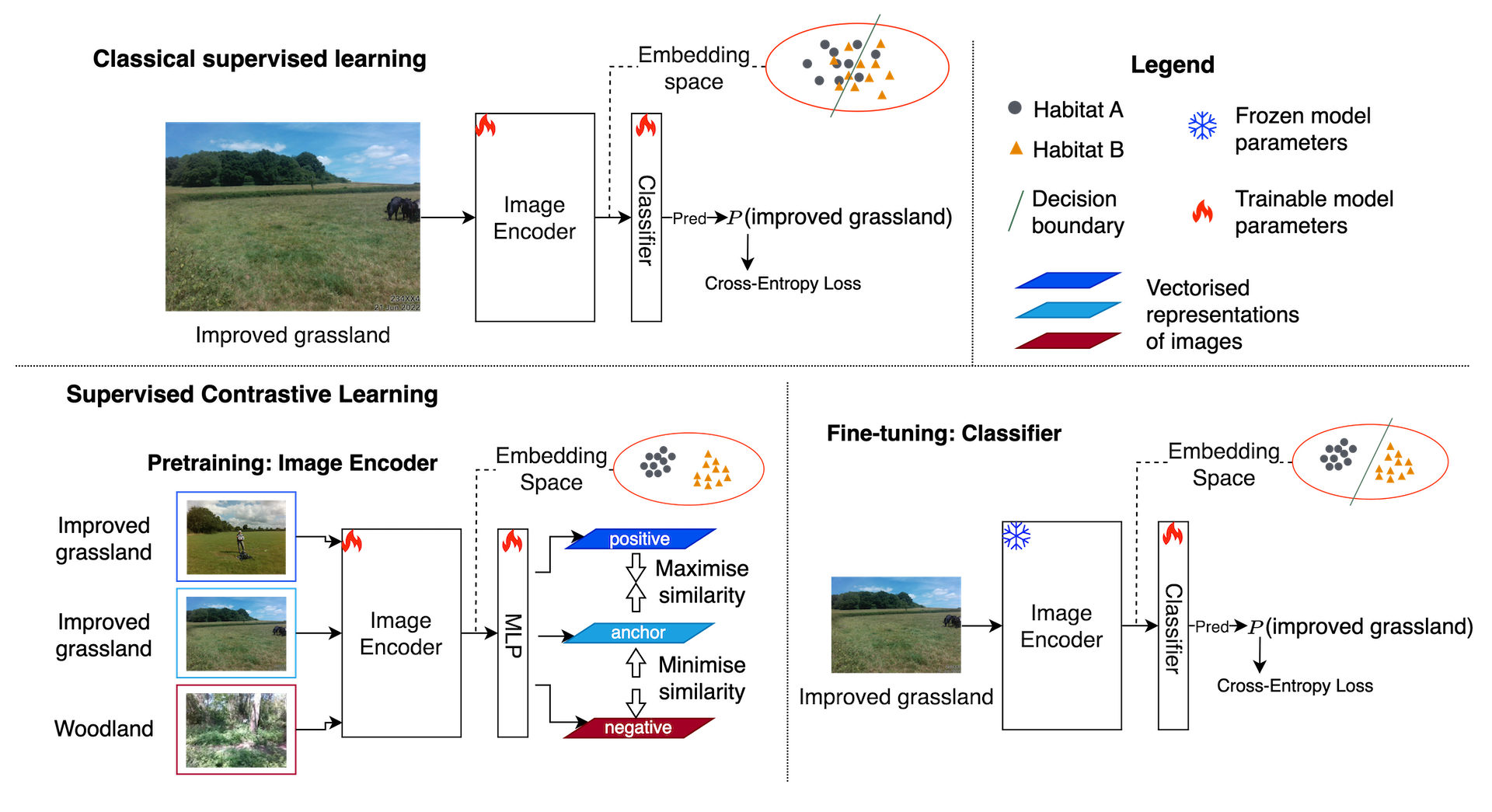}
  \end{center}
  \caption{Comparison of Classical Supervised Learning and Supervised Contrastive Learning. Top (Classical Supervised Learning). An image encoder and classifier are trained end-to-end with cross-entropy. Because embeddings from different classes remain entangled, the classifier struggles to carve out a decision boundary in a poorly separated embedding space, which can hinder performance. Bottom (Supervised Contrastive Learning). First, the encoder is pretrained with a contrastive loss that pulls together same-class embeddings and pushes apart different-class embeddings. This step learns an encoder that generates well-clustered representations. A classifier is then fine-tuned on these fixed embeddings using classic supervised learning, enabling it to learn a much clearer decision boundary. }
  \label{fig: method rationale, sup vs supcon}
\end{figure*}

As the CS dataset is labelled by experts in the field (ground-truthed), we begin by fitting the chosen model via classical supervised learning: minimising the cross-entropy loss between the model's predictions and the ground-truth habitat labels. However, visually similar habitat types pose significant challenges to model optimisation under the classic supervised learning paradigm. Illustrated in Figure \ref{fig: method rationale, sup vs supcon} (a), some habitat classes—such as Neutral Grassland vs. Improved Grassland—share very similar visual patterns that are too subtle to distinguish. When extracting the embeddings of samples from the outputs of the image encoder, which is just ahead of the classifier, and project them into a 2D space, we anticipate to observe significant overlaps between classes (shown as a conceptual illustration in Figure \ref{fig: method rationale, sup vs supcon}). This low inter-class variance in the embedding space makes it difficult to learn a classifier that can carve out a clean decision boundary through the conventional supervised approach. 

To encourage more separable embeddings, we use contrastive learning~\citep{wu2018unsupervised} for a comparison with the conventional supervised learning. Built on the prior knowledge that a model should learn invariant representations for similar samples, contrastive learning pulls together any given training image, denoted as an \textit{anchor} and its augmented views (\textit{positive samples}), while pushing apart all other different images (\textit{negative samples}) in the embedding space. Contrastive learning enables the model to learn meaningful representations for images without references to ground-truth labels. 

When labels are available, contrastive learning can be more flexible for learning representations. Particularly, Supervised Contrastive Learning (SupCon) \citep{khosla2020supervised} integrates the label information into the contrastive loss to force the model to learn close embeddings for instances belonging to the same class, whereas embeddings of different classes are pushed away. As such, SupCon explicitly decreases intra-class variance and boosts inter-class variance, creating tight clusters for each habitat type and clear margins between them. With more separable embeddings, our hypothesis is that SupCon can alleviate the difficulties of learning an effective classifier to discern visually similar habitats.

To implement SupCon, we followed the two-stage transfer-learning strategy proposed in the original SupCon paper: 1) Pretraining: a Multi-Layer Perceptron (MLP) was attached to the image encoder and they were trained together with the SupCon loss on the CS training set. This phase aimed to train the image encoder to represent images into a space where same-class samples cluster and different-class samples repel. 2) Transfer learning: we discarded the MLP, froze the pretrained image encoder, and appended a fresh linear classifier to the image encoder. This classifier was then trained on the CS training set by the standard supervised learning with cross-entropy loss. At inference, we used the image encoder and the classifier trained in the transfer learning stage to predict habitat classes. Figure \ref{fig: method rationale, sup vs supcon} illustrates the training pipelines of classic supervised learning and supervised contrastive learning. 


\subsection{Evaluation} \label{sec: evaluation metrics}
Our models are evaluated with respect to their classification performance and interpretability in predictions, as well as benchmarked against domain specialists.

\subsubsection{Classification Performance}
We employ Top-1 accuracy, Top-3 accuracy, and Matthews Correlation Coefficient (MCC) to evaluate the classification performance on the test set. They are defined as follows:

\begin{equation}
\mathit{Accuracy@1}
= \frac{c^{(1)}}{s}
\end{equation}

\begin{equation}
\mathit{Accuracy@3}
= \frac{c^{(3)}}{s}
\end{equation}

\begin{equation}
\mathit{MCC}
= \frac{c^{(1)}\,s \;-\;\sum_{k} t_k\,p_k}
  {
    \sqrt{s^2 - \sum_{k}p_k^2}\;\sqrt{s^2 - \sum_{k}t_k^2}
  }
\end{equation}

where $c^{(1)}$ is the total number of samples correctly predicted by the top-1 prediction, $c^{(3)}$ is the total number of samples correctly predicted by the top-3 predictions, $s$ is the total number of samples, $t_k$ is the number of times class $k$ truly occurred, and $p_k$ is the number of times class $k$ was predicted (top-1). 

Top-1 accuracy is an evaluation metric that is broadly used to show the percentage of the instances that are correctly classified. Top-3 accuracy is used from a more practical perspective of habitat classification, where the model can still be helpful by producing the top three most possible labels rather than just one. MCC is employed due to the class imbalance of the CS dataset, capturing the model performance while taking into account the highly skewed nature of this dataset (see Figure \ref{fig: CS dataset distributions}). Additionally, we use Precision, Recall, and $F_1$ score to evaluate individual habitat performance. Class-wise, they are defined as follows:

\begin{equation}
\mathit{Precision} = \frac{\textup{TP}}{\textup{TP} + \textup{FP}}
\end{equation}

\begin{equation}
\mathit{Recall} = \frac{\textup{TP}}{\textup{TP} + \textup{FN}}
\end{equation}

\begin{equation}
\mathit{F_1} = \frac{2\,\textup{TP}}{2\,\textup{TP} + \textup{FP} + \textup{FN}}
\end{equation}

where TP is the number of true positives, FN is the number of false negatives, and FP is the number of false positives. We further calculate the weighted $F_1$ score for the overall classification performance while accounting for class imbalance. Weighted $F_1$ calculates the $F_1$ score for each class, then takes a support-weighted average of these per-class values, where “support” is the number of true instances of each class. Moreover, confusion matrix (CM) normalised by the ground-truth labels (columns add to one) is also utilised to illustrate classification performance for each habitat.

\subsubsection{Model Interpretability}

GradCAM \citep{selvaraju2017grad} is utilised for explaining model predictions. GradCAM is a widely accepted technique to generate visual explanations for predictions made by deep neural networks on vision tasks. For image classification tasks, GradCAM produces a class activation map (CAM) using the gradients of the target class to highlight the crucial regions in the image for the model to predict the class. The heated regions shown in warm colours on the map suggests the class is determined by the visual features within these areas, and vice versa. GradCAM can be used to visually examine if model predictions are close to human perceptions. For instance, the GradCAM of a grassland image is expected to highlight regions that are largely occupied by grasses rather than the field surveyor. 

\subsubsection{Evaluation against Domain Specialists} \label{sec: experts specification}

We employ domain specialists to perform a blind annotation exercise on a subset of the test set, aiming to reveal the performance gaps between our model and human experts in evaluating habitat photographs. Specifically, we randomly selected 10\% of the test set, 158 samples in total, for expert review. The draw was stratified to ensure that the subset preserved the distribution of the original test set. Our human specialist benchmark consists of (i) a research scientist with a broad international background and familiarity with UK conservation management, (ii) a research scientist with expertise in long-term land cover dynamics, drivers of vegetation change, habitat management and restoration, and (iii) a senior landscape ecologist who has been working on farmland biodiversity and the UK Countryside Survey for many years. We evaluate each expert's annotations against the ground-truth labels using Top-1 accuracy, Top-3 accuracy, weighted $F_1$ score, and MCC.


\subsection{Implementation Details} \label{sec: method, implementation details}
We constructed the training and test sets by splitting the raw CS dataset into 4200 training instances (75\% of CS dataset) and 1398 test instances (25\% of CS dataset). Applied with stratified partition, the resulting sets preserve approximately the same class distribution as the complete CS dataset. As the habitats of Supra-littoral Rock and Littoral Rock have too few samples in the CS dataset, the stratified partition did not include them in the test set. Additionally, 20\% of instances in the training set were used for validating hyperparameter choices with the grid search strategy. 

Our implementation was based on the PyTorch library. The AdamW~\citep{loshchilov2017decoupled} optimiser was chosen for model training.  The learning rate, weight decay, and the batch size were tuned to $5 \times 10^{-6}$, 0.05, and 16 respectively. With respect to the data preprocessing pipeline during training, images were first resized to 384 $\times$ 384 and then applied with random crop and rotation transformation before being fed to the model. The model was trained for 50 epochs as the learning performance was saturated by this point. In terms of the hardware, all of our experiments were run on a single NVIDIA 4090 GPU with 24GB memory. 

For hyperparameters unique to SupCon, we followed its original settings proposed in \cite{khosla2020supervised}. During the pretraining stage, the projection head used was a two-layer MLP with the ReLU activation. The batch size was doubled by independently applying data transformations of random crop and rotation for two times. The temperature in the contrastive loss was set to 0.1. Contrastive pretraining was conducted for 100 epochs. However, we found the transfer performance using the pretrained image encoder plateaued after 50 epochs. Accordingly, the image encoder pretrained for 50 epochs was adopted for the transfer learning experiments.

%% file: experiments.tex
\section{Results}
Based on our model architecture choices for habitat classification presented in Section \ref{sec: models}, we first report and discuss the performance of CNN baselines and ViTs in this section, examining both classification performance and model interpretability. Subsequently, we compare SupCon with classical supervised learning described in Section \ref{sec: training methods}, focusing on classification performance on challenging habitats that are visually similar. Finally, we evaluate our best-performing model against human experts on a subset of the test data, revealing the performance gap and discussing the potential of our approach for real-world deployment. 

\subsection{Classification Performance} \label{sec: method, pretraining}


Table~\ref{tab: SwinT vs. CNNs.} compares the classification performance of the chosen CNNs and the ViTs on the habitat classification of the test set. SwinT models outperform all CNN baselines by noticeable margins in all four classification metrics. Specifically, SwinT-B improves Top-1 accuracy, $F_1$ score, and MCC by 3.86\%, 0.046, and 0.043, respectively. Top-3 accuracy is increased by 3.08\% with the SwinT-L, exceeding 91\%. In terms of the performance of CNN baselines, two WRNs, EfficientNetV2-L, and ResNeXt-50 are on par with each other, with less than 0.6\% performance difference on Top-1 accuracy which could be attributed to the noise during training. The close performance observed in the CNNs suggests that this type of model could have reached a performance ceiling on the habitat classification task, which is surpassed by SwinT models.

\begin{table*}
\caption{Classification performance: CNNs vs. SwinT. Top-1, Top-3 accuracies, weighted $F_1$ score and MCC are used to evaluate the classification performance of models on the test set of CS data. Bold numbers indicate best performance and increments, while underlined numbers are best baselines for comparisons.}
\centering
\begin{tabular}{@{}llllll@{}}
\toprule
\textbf{Model} & \textbf{\# Para. (M)} & \textbf{Top-1 Acc. (\%)}   & \textbf{Top-3 Acc. (\%)} & \textbf{$F_1$ (weighted)} & \textbf{MCC} \\ \midrule
WRN-50-2 & 66.88 & 65.59 & 87.63 & 0.640 & 0.608  \\ 
WRN-101-2 & 124.88 & 65.52 & 86.98 & \underline{0.644} & 0.608  \\ 
EfficientNetV2-M & 52.88 & 63.38 & 85.77 & 0.623 & 0.585   \\ 
EfficientNetV2-L & 117.26 & 65.95 & \underline{88.34} & 0.642 & 0.613   \\ 
ResNeXt-50 & 23.88 & \underline{66.17} & 86.91 & 0.642 & \underline{0.616}   \\ 
ResNeXt-101 & 81.45 & 64.45 & 88.27 & 0.623 & 0.596   \\ \midrule
SwinT-S & 48.85 &  67.53 & 90.27 & 0.658 &  0.632 \\
SwinT-B & 86.90 &  \textbf{70.03 (+3.86)} & 90.70  & \textbf{0.690 (+0.046)} & \textbf{0.659 (+0.043)}\\
SwinT-L & 195.23 &  69.74 & \textbf{91.42 (+3.08)} & 0.688 & 0.656\\

\bottomrule
\end{tabular}
\label{tab: SwinT vs. CNNs.}
\end{table*}

In addition to the overall performance, we further make detailed pairwise comparison between representatives of the two model architectures on individual habitats. Particularly, ResNeXt-50 and SwinT-B are employed as the representatives. ResNeXt-50 is a strong CNN baseline that achieves the best performance on two out of four classification metrics, whereas SwinT-B outperforms SwinT-S and SwinT-L on three classification metrics. Table~\ref{tab: individual_class_performance} compares SwinT-B with ResNeXt-50 by breaking down the weighted $F_1$ score shown in Table~\ref{tab: SwinT vs. CNNs.} into precision, recall, and $F_1$ score on major habitat types. To simplify the results, we grouped L3 performance into L2 categories. For instance, Improved Grassland and Neutral Grassland both belong to the grassland group and each prediction from them contributes to the precision, recall, and the resulting $F_1$ score of the grassland group. In summary, Table~\ref{tab: individual_class_performance} reveals that the SwinT-B surpasses ResNeXt-50 across all individual habitats on $F_1$ score, ranging between 0.004 to 0.112.  

\begin{table*}[h]
    \centering
    \caption{Comparing SwinT-B vs. ResNeXt-50 on individual habitats. This table shows the classification performance when L3 habitats are aggregated into corresponding L2 habitats (to simplify the results presented). Precision and recall are shown along with the $F_1$ score. Bracketed values represent the performance change of SwinT-B relative to ResNeXt-50.}
    \label{tab: individual_class_performance}
    \begin{tabular}{l l l c c c}
        \toprule
        \textbf{Coarse Class} & \textbf{\#Test Samples} & \textbf{Model} & \textbf{Precision} & \textbf{Recall} & \textbf{$F_1$} \\ 
        \midrule
        \multirow{2}{*}{Grasslands} 
            & \multirow{2}{*}{579} 
            & ResNeXt-50   & \textbf{0.601}  & 0.603  & 0.602 \\
            &                      & SwinT-B  & 0.585 (-0.018)  & \textbf{0.629 (+0.027)}  & \textbf{0.606 (+0.004)} \\ \midrule
        \multirow{2}{*}{Croplands}  
            & \multirow{2}{*}{240} 
            & ResNeXt-50   & 0.807  & \textbf{0.942}  & 0.869 \\
            &                      & SwinT-B  & \textbf{0.875 (+0.068)}  & 0.929 (-0.013)  & \textbf{0.901 (+0.032)} \\ \midrule
        \multirow{2}{*}{Wetlands}   
            & \multirow{2}{*}{226} 
            & ResNeXt-50   & 0.590  & \textbf{0.637}  & 0.613 \\
            &                      & SwinT-B  & \textbf{0.782 (+0.192)}  & 0.620 (-0.017)  & \textbf{0.691 (+0.045)} \\ \midrule
        \multirow{2}{*}{Woodlands}  
            & \multirow{2}{*}{200} 
            & ResNeXt-50   & 0.789  & 0.655  & 0.716 \\
            &                      & SwinT-B  & \textbf{0.801 (+0.012)}  & \textbf{0.745 (+0.090)}  & \textbf{0.772 (+0.056)} \\ \midrule
        \multirow{2}{*}{Heathlands} 
            & \multirow{2}{*}{123} 
            & ResNeXt-50   & 0.580  & 0.561  & 0.570 \\
            &                      & SwinT-B  & \textbf{0.645 (+0.065)}  & \textbf{0.724 (+0.163)} & \textbf{0.682 (+0.112)} \\ 
        \bottomrule
    \end{tabular}
\end{table*}

\subsection{Prediction Interpretability}

\begin{figure*}
  \begin{center}
    \includegraphics[width=0.95\linewidth]{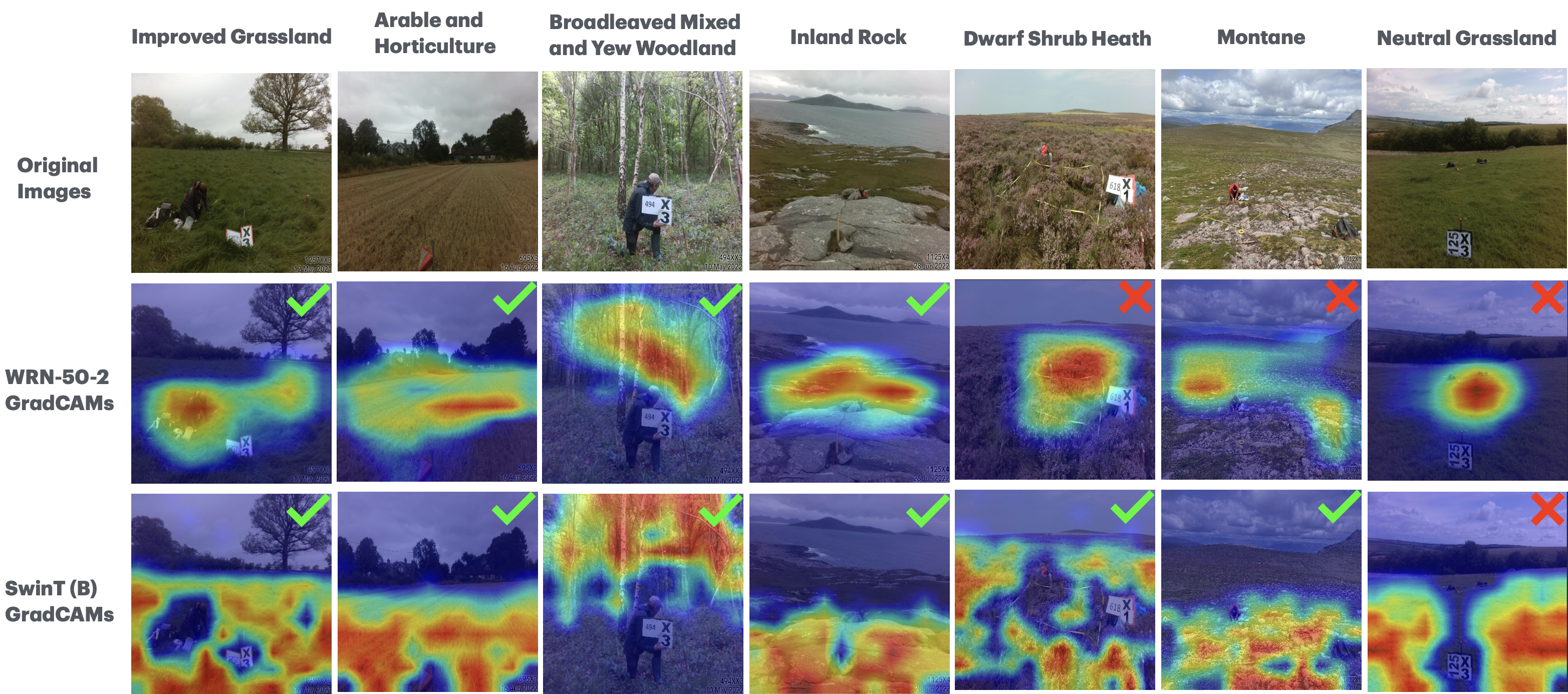}
  \end{center}
  \caption{Comparison of the GradCAMs generated by WRN-50-2 and SwinT-B. The green notation suggests a correct prediction on the sample and the red notation indicates a misclassification. Whether correct or not, the SwinT-B judges the habitat based on a broader and more connected background across the entire image, with a particular overlook on irrelevant visual features, resulting in an improved interpretability over WRN-50-2 on habitat classification task. }
  \label{fig: gradcams}
\end{figure*}

The merits of employing SwinT models on our habitat classification task are apparent from the results shown in Tables \ref{tab: SwinT vs. CNNs.} and \ref{tab: individual_class_performance}. Furthermore, SwinT exhibits superiority on the interpretability over CNNs by paying attention to broader visual features than its CNN counterparts on habitat images.  

Specifically, we demonstrate the model interpretability with GradCAMs. Figure~\ref{fig: gradcams} displays exemplary GradCAMs produced by SwinT-B and WRN-50-2 on different habitats. The observations are twofold: 1) Compared to WRN-50-2, the prediction of SwinT-B is based on a much larger visual context of the images. In contrast, the visual features leveraged by the WRN-50-2 for inference is rather narrow-scoped. The classification made by SwinT-B is much more intuitive as its broad focus is more suitable for making judgement for habitat images. 2) SwinT-B largely ignores distracting visual features, such as on-field surveyors and their equipment, which are irrelevant for habitat classification, whereas, WRN-50-2 is occasionally distracted by such visual features (see the first column and the last column in Figure~\ref{fig: gradcams}). Similar GradCAMs to WRN-50-2 are found on other employed CNN models in this study, indicating CNNs could correlate non-determinist visual objects to its predictions, subsequently leading to poor interpretability. 

We attribute the illustrated interpretability of SwinT-B for habitat classification to its attention mechanism. Take the grassland as an example, SwinT learns to connect small grass regions shown across the entire image as they are visually similar to each other, and leverage the learned embeddings of connected grass regions to activate its classification layer for the prediction. Conversely, the prediction of CNN is based on confined local areas, without a learning mechanism to connect them over the entire image, resulting in its subjectiveness to local visual features, which makes CNN a poor fit for habitat classification. Combining the observations in classification performance and GradCAMs, we conclude that the capability of ViT for learning image representation encoding long-range regional dependencies and the overall contextual information of the scene, is crucial for improving the model for habitat recognition, and results in superior classification performance over CNN models. While SwinT generalises over a broad scene context, this does not necessarily lead to correct predictions. As shown in Figure \ref{fig: gradcams} (rightmost column), SwinT misclassifies a Neutral Grassland photo to Improved Grassland. The error arises from the difficulty of distinguishing habitats with highly similar visual context. Neutral and Improved Grasslands are often subtly different and difficult to interpret in the field. Hence, addressing confusions caused by similar visual context is essential to further improve overall accuracy.


\subsection{Comparing SupCon with Supervised Learning}

This section compares SupCon with classical supervised learning for training the model with the best reported performance in the previous section, the SwinT-B, and aims to validate the hypothesis that SupCon can improve the classification performance on visually similar habitats. 

\begin{figure*}
  \begin{center}
    \includegraphics[width=0.95\linewidth]{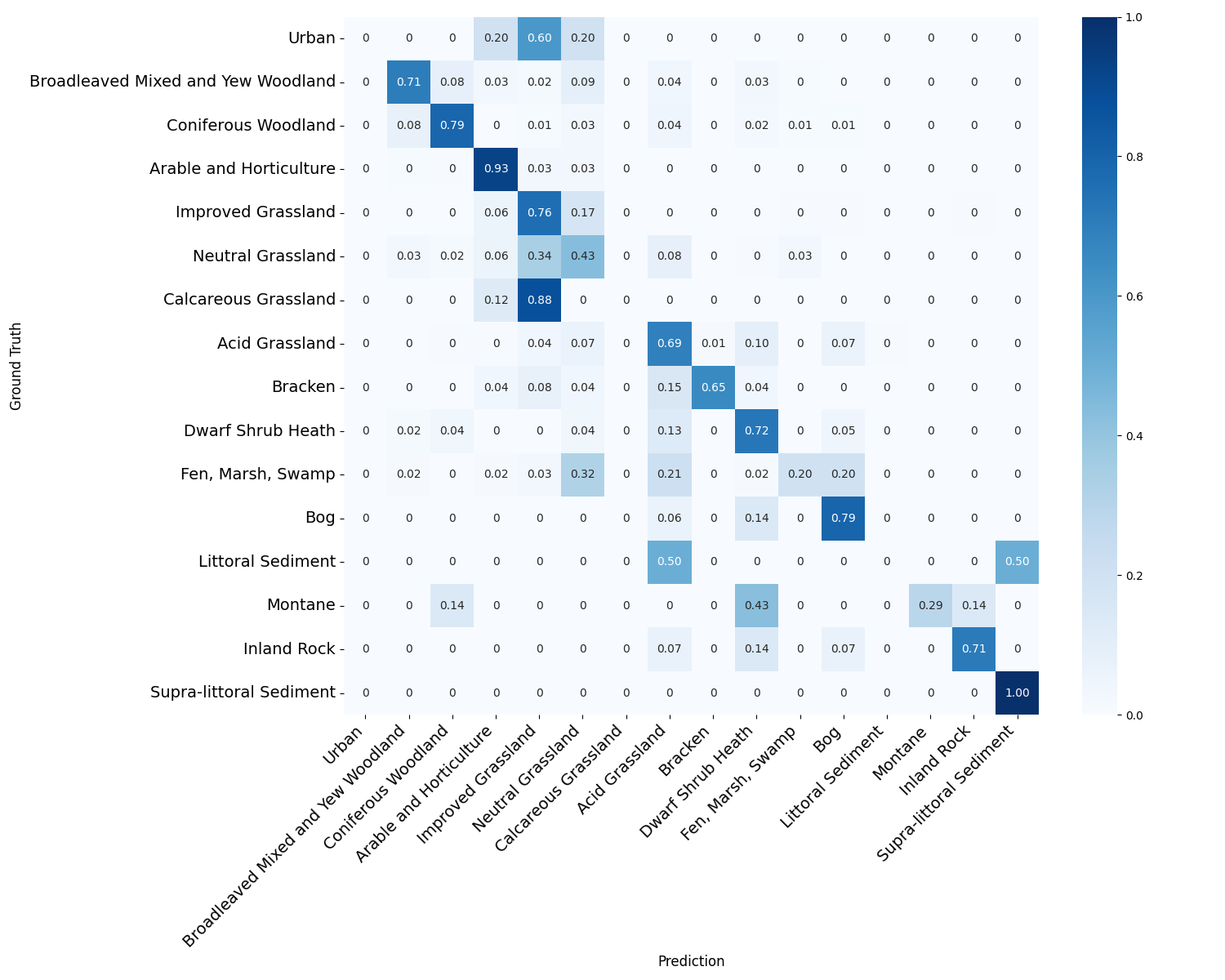}
  \end{center}
  \caption{The confusion matrix of SwinT-B on the test set. Habitats with no samples in the test set are removed from the matrix (refer to Section \ref{sec: method, implementation details} for details). While SwinT-B demonstrates more suitability than CNNs for habitat classification, some habitats are often misclassified by the model, e.g., Neutral Grassland is often misclassified as Improved Grassland, affecting the overall classification performance significantly.}
  \label{fig: swint cm full}
\end{figure*}

Table~\ref{tab: SwinT vs. CNNs.} has reported the classification performance achieved by SwinT-B under the supervised learning paradigm. Figure~\ref{fig: swint cm full} further shows the classification performance on individual L3 habitats achieved by the supervised learning by illustrating the normalised confusion matrix (CM) produced by SwinT-B. For major habitats, it is observed that the classification performance is relatively modest on grasslands, such as Neutral Grassland, Acid Grassland, and Bracken, as well as on Fen, Marsh, Swamp (FMS) belonging to the wetland. Less than 70\% of test instances among these habitats are correctly predicted by SwinT-B (shown by the diagonal values in the corresponding rows of CM). This observation indicates that these habitats are very challenging to classify even using ViT (CNNs perform closely to ViT or even worse on these habitats in our experiments). We attribute this performance degradation to the visual similarity between certain habitats, as illustrated in Figure \ref{fig: intro, all habitats}. For instance, Neutral Grassland, which is a major L3 habitat in the CS dataset and often displays subtle visual differences to Improved Grassland, has 34\% of its test instances being misclassified as Improved Grassland, deteriorating the overall performance significantly. In Appendix A, we visualise the embedding clusters of different habitats. The heavily overlapped clusters further reflect their high similarity.

\begin{table}[t]
\centering
\caption{Performance of supervised learning (Sup) vs.\ supervised contrastive learning (SupCon) across architectures. 
Numbers are reported on the CS habitat test set. For each model, SupCon cells show the absolute gain over Sup in parentheses.}
\label{tab: sup-vs-supcon-all}
\setlength{\tabcolsep}{6pt}
\begin{tabular}{lcccc}
\toprule
\textbf{Model}  & \textbf{Top-1 Acc. (\%)} & \textbf{Top-3 Acc. (\%)} & $\mathbf{F_{1}}$ \textbf{(weighted)} & \textbf{MCC}\\
\midrule
\textbf{WRN-50-2}  & 64.23 \,(\,-1.36) $\downarrow$ & 89.13 \,(+1.50) $\uparrow$ & 0.627 \,(\,-0.013) $\downarrow$ & 0.593 \,(\,-0.015) $\downarrow$\\
\addlinespace[2pt]
\textbf{EfficientNetV2-L}  & 67.02 \,(\,+1.07) $\uparrow$ & 90.70 \,(\,+2.36) $\uparrow$ & 0.658 \,(\,+0.016) $\uparrow$ & 0.625 \,(\,+0.012) $\uparrow$ \\
\addlinespace[2pt]
\textbf{ResNeXt-50} & 64.23 \,(\,-1.94) $\downarrow$ & 88.20 \,(\,+1.29) $\uparrow$ & 0.610 \,(\,-0.032) $\downarrow$ & 0.593 \,(\,-0.023) $\downarrow$ \\
\addlinespace[2pt]
\textbf{SwinT-B}  & \textbf{71.53} \,(\,+1.50) $\uparrow$ & \textbf{90.92} \,(\,+0.22) $\uparrow$ & \textbf{0.706} \,(\,+0.016) $\uparrow$ & \textbf{0.677} \,(\,+0.018) $\uparrow$ \\
\bottomrule
\end{tabular}
\end{table}


\begin{figure*}
  \begin{center}
    \includegraphics[width=0.99\linewidth]{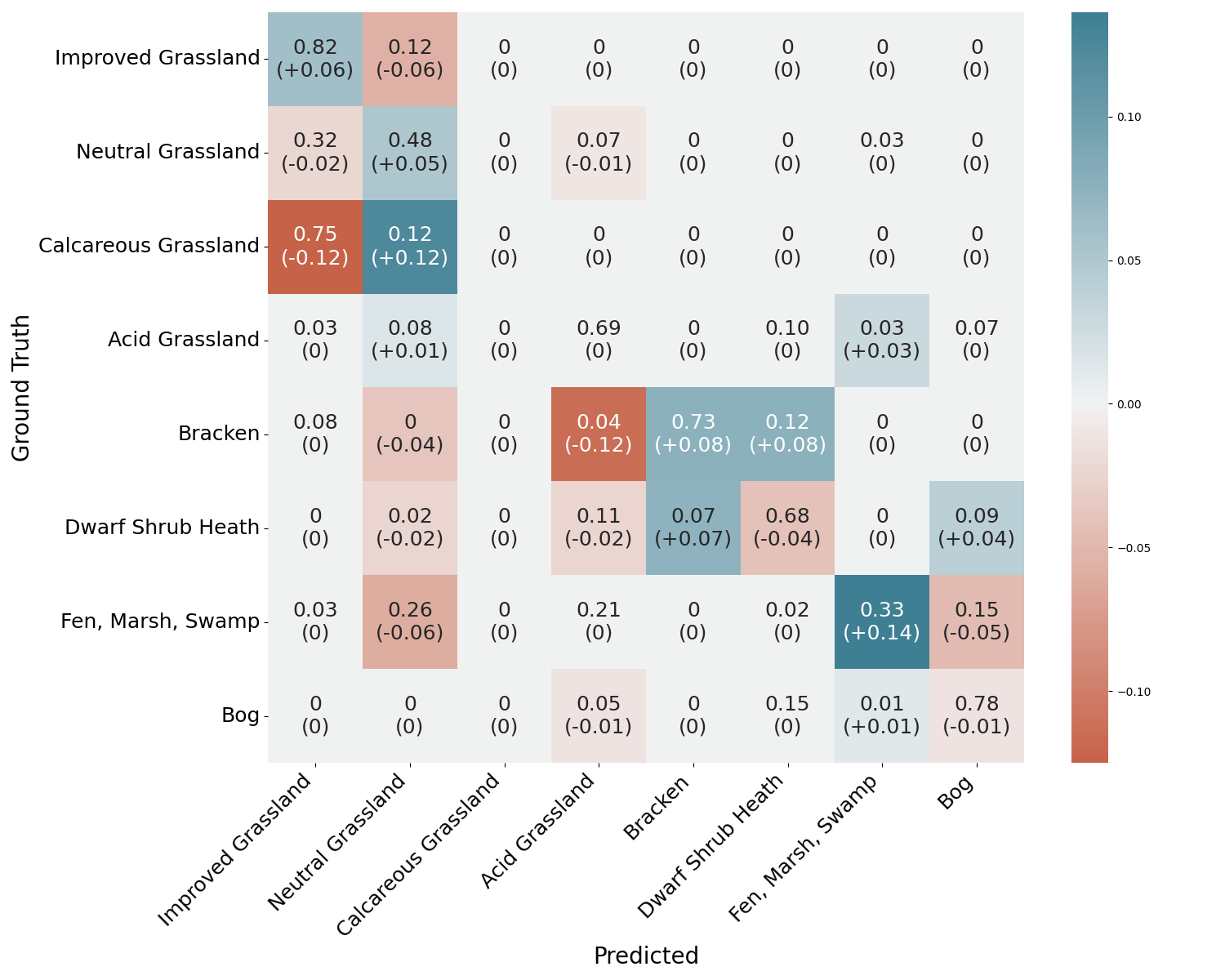}
  \end{center}
  \caption{Delta confusion matrix (CM) of challenging habitats: based on the CM produced by SupCon, this graph further highlights its differences from the CM generated by supervised learning, with blue colour indicating boost and red colour indicating decrease. SupCon reduces misclassifications on three major grasslands and the FMS in wetlands, which confuse the model most in the supervised learning.}
  \label{fig: sup vs supcon cm}
\end{figure*}

We applied SupCon to the top-performing variant from each CNN type (WRN-50-2, EfficientNetV2-L, ResNeXt-50) and to the strongest Swin Transformer (SwinT-B), as identified in our supervised learning comparison, using identical contrastive pretraining settings. Table~\ref{tab: sup-vs-supcon-all} reports their classification performance, denoted with the changes based on the classical supervised learning indicated in Table~\ref{tab: SwinT vs. CNNs.}. SupCon slightly improves the performance of SwinT-B on all four metrics (+1.50 \% Top-1, +0.22 \% Top-3, +0.016 $F_1$, +0.018 MCC). Regarding CNNs, SupCon gives marginal gains for EfficientNetV2-L across metrics and has mixed effects on WRN-50-2 and ResNeXt-50 with slight improvements in Top-3 but degradations in Top-1/$F_1$/MCC. These results suggest that, SupCon provides the clearest benefit for the Swin Transformer under the same settings, with inconsistent performance for the CNN baselines.

For our best model, SwinT-B, SupCon is observed to mitigate the misclassifications between challenging habitats. Specifically, Figure~\ref{fig: sup vs supcon cm} is a delta confusion matrix that compares the normalised confusion matrices generated by SupCon and supervised learning on major habitats. With SupCon, Improved Grassland increases its correct classifications by 6\% and reduces the misclassifications on Neutral Grassland by 6\%. Similarly, 5\% more Neutral Grassland instances are correctly predicted and its misclassifications to Improved Grassland are reduced by 2\% with SupCon. In terms of FMS, SupCon improves the proportion of correct classifications by 14\% and decrease its misclassifications to Bog and Neutral Grassland by 5\% and 6\%, respectively. The correct predictions of Bracken are improved by 8\% with SupCon, owing to a 12\% decrease of misclassifications to Acid Grassland. Conversely, SupCon sacrifices classification performance of Dwarf Shrub Heath, decreasing its correct predictions by 4\%, and fails to increase the correct classifications for Calcareous Grassland. This could be due to the fact that the Calcareous Grassland is a minor L3 habitat with a very limited number of samples in CS dataset for training the model, as shown in Figure~\ref{fig: CS dataset distributions}. We provide more insights about the performance of SupCon with the embedding analysis in Appendix A.

For our CNN backbones, we observe that SupCon can shift the decision boundary for them rather than enlarge the inter-class margin: for Neutral vs Improved Grassland, SupCon reduces Neutral to Improved confusions but increases Improved to Neutral, yielding higher recall for Neutral at the expense of precision. A plausible explanation is the strong local prior of CNNs, which may limit the visual information expressed in the embeddings, making contrastive learning less effective to separate classes that require broader contextual cues. In contrast, SwinT-B with SupCon reduces errors in both directions, suggesting that global contextual cues leveraged by the SwinT-B are better suited to contrastive learning for the habitat classification task. Further details on the performance of CNNs with SupCon are provided in the Appendix B.

\subsection{Comparing the Model with Human Experts}

\begin{table}[ht]
\centering
\caption{Performance comparison between our best models, EfficientNetV2-L and SwinT-B, and human experts on a subset drawn from the test set. Bold values are the best performance, underlined values are the best performance achieved by human experts, and performance differences between the model and the best human performance are indicated in the brackets.}
\label{tab: model_vs_expert}
\begin{tabular}{@{}lcc|ccc@{}}
\toprule
\textbf{Metric}     &    \textbf{EfficientNetV2-L (SupCon)}          & \textbf{SwinT-B (SupCon)} & \textbf{Expert 1} & \textbf{Expert 2} & \textbf{Expert 3}  \\ 
\midrule
Top-1 Acc. (\%)   & 56.33 (-7.36)  & \textbf{63.92 (+0.23)}        & 55.70     & 51.27 &\underline{63.69}  \\
Top-3 Acc. (\%)   & 84.18 (+4.56) & \textbf{87.34 (+7.72)}        & 73.42     & 74.68 & \underline{79.62} \\
$F_1$ (weighted)   &   0.535 (-0.093)   & 0.603 (-0.025)        & 0.541     &  0.529 & \underline{\textbf{0.628}} \\
MCC vs Ground Truth  &   0.516 (-0.088) & 0.599 (-0.005)       & 0.517     &  0.479 & \underline{\textbf{0.604}} \\
MCC vs Expert 1  & 0.529  & 0.529  & -- & -- & --         \\
MCC vs Expert 2  & 0.460  & 0.490  & 0.515 & -- & --      \\
MCC vs Expert 3  &  0.530  & 0.546  & 0.511 & 0.553  & --      \\
\bottomrule
\end{tabular}
\end{table}

Table~\ref{tab: model_vs_expert} compares the performance of our best performed CNN and ViT trained with SupCon as shown in Table~\ref{tab: sup-vs-supcon-all}, on a subset of the test data with photographic interpretation of L3 habitats by three human experts, as specified in Section \ref{sec: experts specification}. The notably higher performance of Expert 3 likely reflects their extensive familiarity with the test data, gained through years of participation in the UK Countryside Survey. Compared to the best expert (Expert 3), SwinT-B: 1) outperforms their Top-1 accuracy by 0.23\% and their Top-3 accuracy significantly by 7.72\%; 2) slightly underperforms on weighted $F_1$ and MCC by 0.025 and 0.005, respectively, primarily because it struggles to generalise to rare habitats (e.g. Montane) owing to limited training samples (see Appendix C for detailed discussion); 3) has a strong agreement with the best expert, achieving an MCC of 0.546 against them. In contrast, our best CNN with SupCon (EfficientNetV2-L) underperforms the human experts on all headline metrics and is well below SwinT-B, indicating that studied CNNs do not yet reach expert-level performance on this task.

\section{Discussion}

Our experiment results show that SwinT surpasses state-of-the-art CNNs, including Wide ResNet, EfficientNetV2, and ResNeXt, for ground-level habitat classification. SupCon training can further boost SwinT, achieving performance comparable to human experts. Meanwhile, the study also exposes challenges in habitat classification with ground-level imagery. We provide the following insights: 

\begin{itemize}
    
    \item \textbf{Inherent challenge of ground-level imagery.} With a Top-1 accuracy of only \textasciitilde 63\% in human benchmark, our study reveals that ground-level classification of L3 habitats from images remains difficult for both AI models and human experts. Remote sensing methods have shown better performance in multi‐habitat (albeit narrower) contexts, e.g. forests \citep{pratico2021machine}, wetlands \citep{jamali2023wetmapformer}, grasslands \citep{diaz2024classification}. Unlike bird’s-eye remote-sensing data, ground-level images require models to understand scene depth (disentangling background from foreground), to filter out distracting elements present in the scene, and to detect often‐subtle compositional differences among visually similar subtypes. These added complexities make ground-level habitat classification very challenging. Misclassification by the AI model can trigger inappropriate actions on the ground. From an ecological perspective, misclassified habitats could lead to wrong interventions or recommendations for land management, e.g. applying mineral fertiliser to neutral/species-rich grassland. For Biodiversity Net Gain, inaccurate mapping could result in development on ecologically valuable habitats or inappropriate decisions about where to offset habitat losses following development. Therefore, addressing the challenges of ground-level image classification is critical to responsible habitat monitoring. 
    
    
    \item \textbf{Pathways to improving ground-level habitat classification.} a) The CS photos were originally taken for plot relocation, not habitat identification. Establishing clear guidelines (e.g. framing, lighting, seasonal timing) for habitat‐focused imagery should enhance the training data and reveal the true potential of AI for ground‐level habitat classification. b) Human surveyors use a rich collection of information and often quantitative data---information absent from 2D images---to determine habitats. For example, they take into consideration the geographical and climatic information, management context, soil type, species, etc. when conducting habitat assessment in the field. Additionally, they may visit the site during different seasons in order to record species growing at different times of the year. Incorporating multi-view or time-series photos, or auxiliary metadata (e.g. date, location), could narrow the remaining performance gap. c) Since habitat is often defined by characteristic species, extending our approach to recognise key flora (e.g. dominant shrubs, indicator plants) and integrating those outputs into the habitat classifier may also boost accuracy. 
    
    \item \textbf{Practical deployment.} When deploying the model in practice, a well-designed sampling protocol is critical to ensure reliable performance. We propose: capturing clear, context-rich images with the rear camera supported by automatic quality checks; sampling multiple photo-points within each spatial unit; and acquiring several views per point (forward, nadir, and detail). Predictions are then aggregated to habitat classes with uncertainty estimates and flags for expert review, ensuring responsible use. From a user-experience perspective, network latency is the principal constraint: a reliable connection is required to transmit images and return results with minimal delay. By contrast, compute time is comparatively negligible: on our workstation (NVIDIA RTX 4090), a single SwinT-B forward pass takes milliseconds. In our web-based API prototype, end-to-end latency is typically 0.5–2.0 s per image. 
    
\end{itemize}

%% file: conclusion.tex
\section{Conclusion}
In this work, we investigated deep learning approaches for habitat classification based on ground-level imagery to bridge the research gaps in AI-driven habitat monitoring. Guided by the unique challenges of ground habitat scenes (broad and continuous regions, visually similar habitats), we compared two deep neural architectures, CNNs and ViTs, and two training paradigms, standard supervised learning vs supervised contrastive learning (SupCon). 

Our evaluation shows that ViTs outperform state-of-the-art CNN baselines—achieving up to a 4\% increase in overall accuracy and similar gains across other key metrics. Beyond raw accuracy, ViTs generalise with a broader habitat scene understanding, demonstrating superior interpretability over CNNs. We also find that SupCon is a more effective training paradigm than classical supervised learning for training ViT to distinguish between visually similar habitats. SupCon considerably reduces misclassifications between commonly confused habitat pairs—such as Improved and Neutral Grassland—outperforming standard supervised learning. Finally, our best ViT model not only exceeds CNN performance but also matches photographic identification by experienced ecological experts on key classification metrics, highlighting the potential of our work to reduce the need for human experts to interpret habitats at this level in the field.


Our current study focuses solely on level 3 Broad Habitats within the UK (except for Bracken, which is L4). Future work should expand the model to classify fine-grained level 4 habitats with over 80 categories, where sample sizes are still small, and assess model transferability to other countries and different habitats outside of the UK. In summary, our study paves a path towards truly scalable, cost-effective habitat monitoring based on ground-level imagery by integrating state-of-the-art AI techniques with ecological expertise to support better-informed biodiversity conservation and land-use decisions.

%% file: appendix.tex
\setcounter{figure}{0}
\renewcommand{\thefigure}{\arabic{figure}}

\setcounter{table}{0}
\renewcommand{\thetable}{\arabic{table}}

\appendix

\section{Embedding Analysis} \label{appendix embedding analysis}

We employ UMAP, CHI, and DBI to analyse the embedding space produced by the trained image encoder, both qualitatively and quantitatively. UMAP is applied to compress the embeddings extracted from the encoder and visualise them in a 2D graph. Embeddings from different classes can be clustered on the UMAP, showing whether they are heavily overlapped or distinctively separated, patterns that can impact the classification performance. UMAP is particularly useful for us to observing the effect of the chosen learning methods, the conventional supervised learning and supervised contrastive learning. In addition, we further utilise the Calinski–Harabasz (CH) index \citep{Caliński01011974}, and Davies–Bouldin (DB) index \citep{4766909} to quantitatively evaluate the inter-class variance and intra-class variance in the embedding space. CH index and DB index are both popular metrics for evaluating clustering performance, rewarding methods that are capable of achieving more separable clusters for features. A higher CH index or a lower DBI implies a better clustering quality, with low intra-class variances and high inter-class variances.

\begin{figure}[htbp]
  \centering
  
    \begin{subfigure}{.48\linewidth}
    \centering
    \centerline{\includegraphics[height=4.5cm]{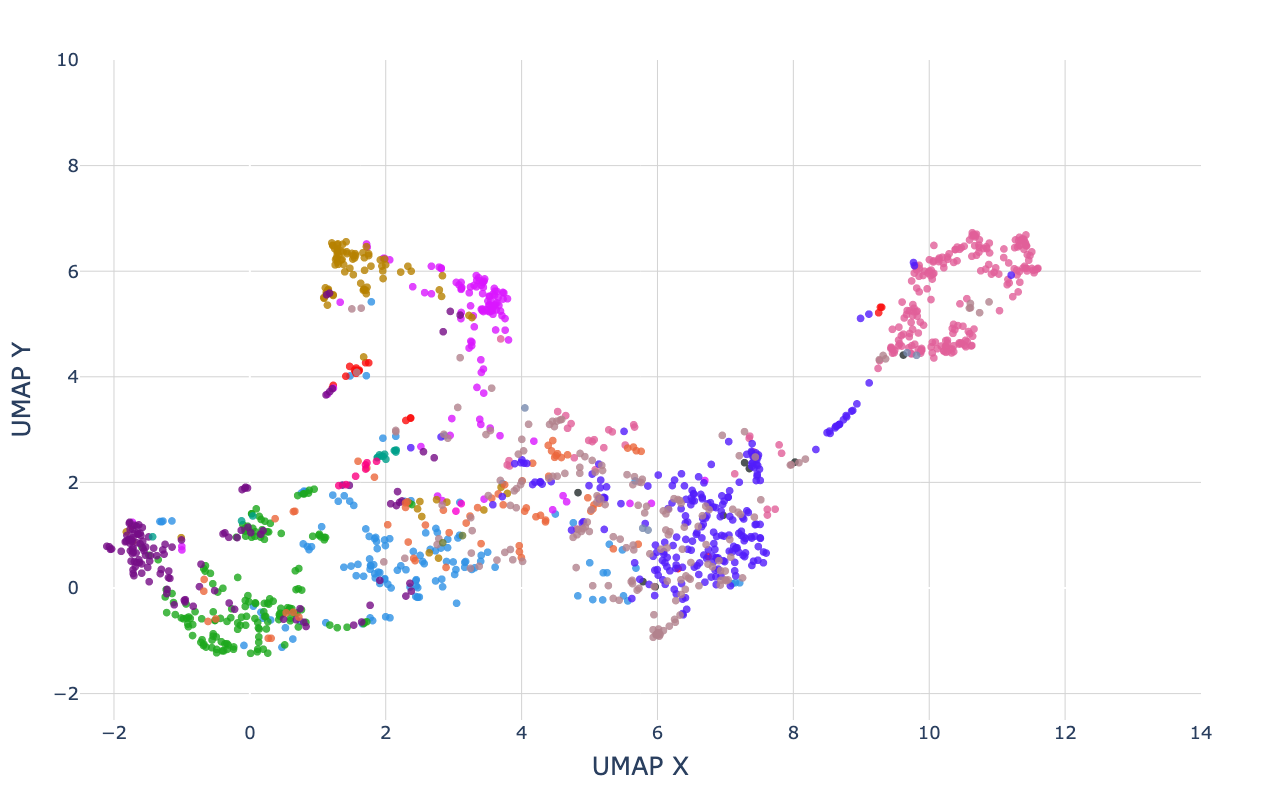}}
    \caption{Sup All Classes}
    \end{subfigure} 
    \begin{subfigure}{.48\linewidth}
    \centering
    \centerline{\includegraphics[height=4.5cm]{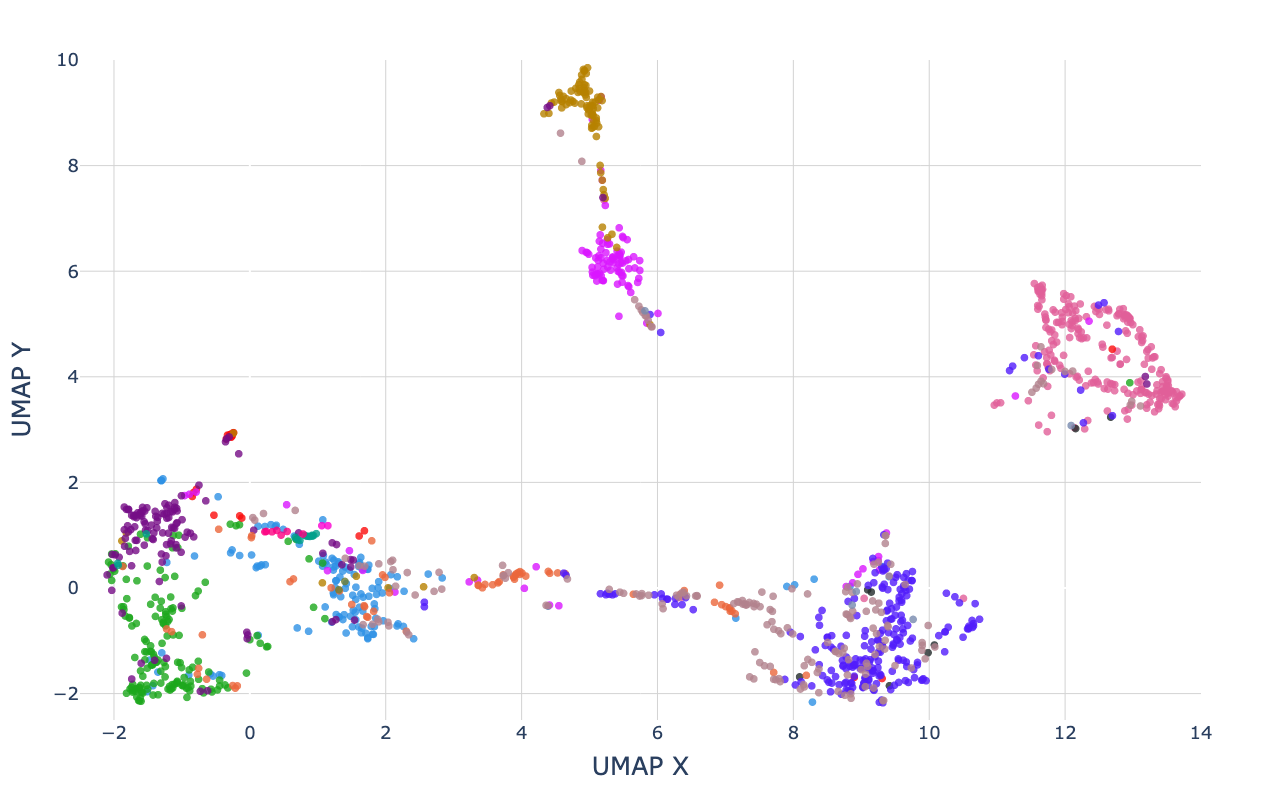}}
    \caption{SupCon All Classes}
    \end{subfigure}

    \begin{subfigure}{.48\linewidth}
    \centering
    \centerline{\includegraphics[height=4.5cm]{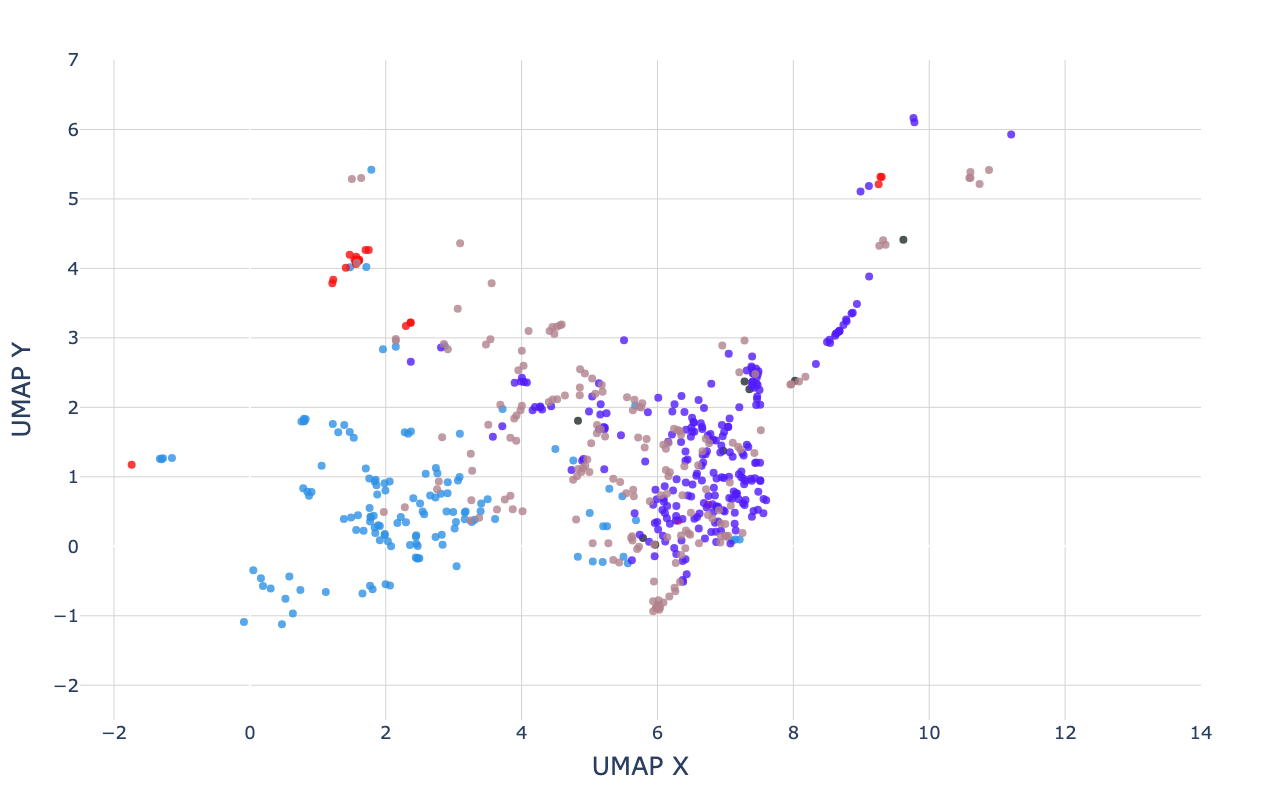}}
    \caption{Sup Grassland}
    \end{subfigure} 
    \begin{subfigure}{.48\linewidth}
    \centering
    \centerline{\includegraphics[height=4.5cm]{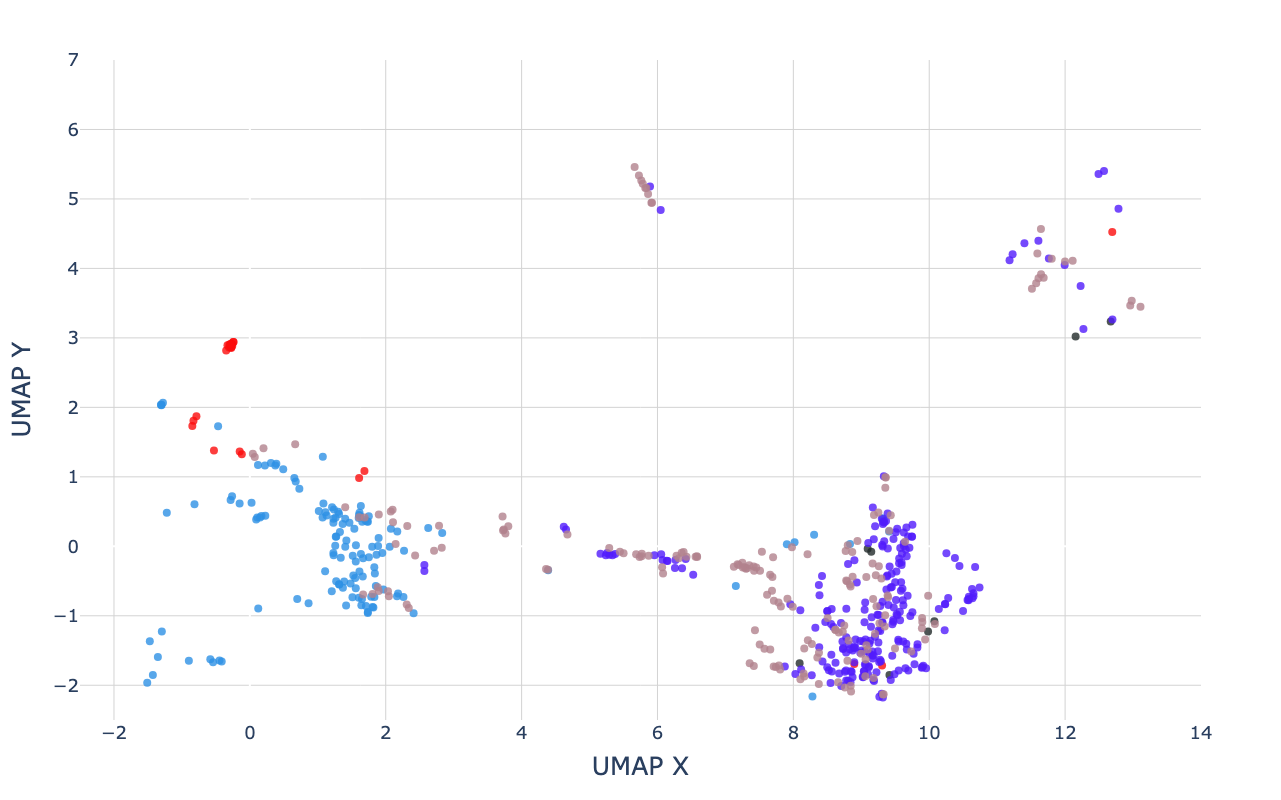}}
    \caption{SupCon Grassland}
    \end{subfigure}

    \begin{subfigure}{.48\linewidth}
    \centering
    \centerline{\includegraphics[height=4.5cm]{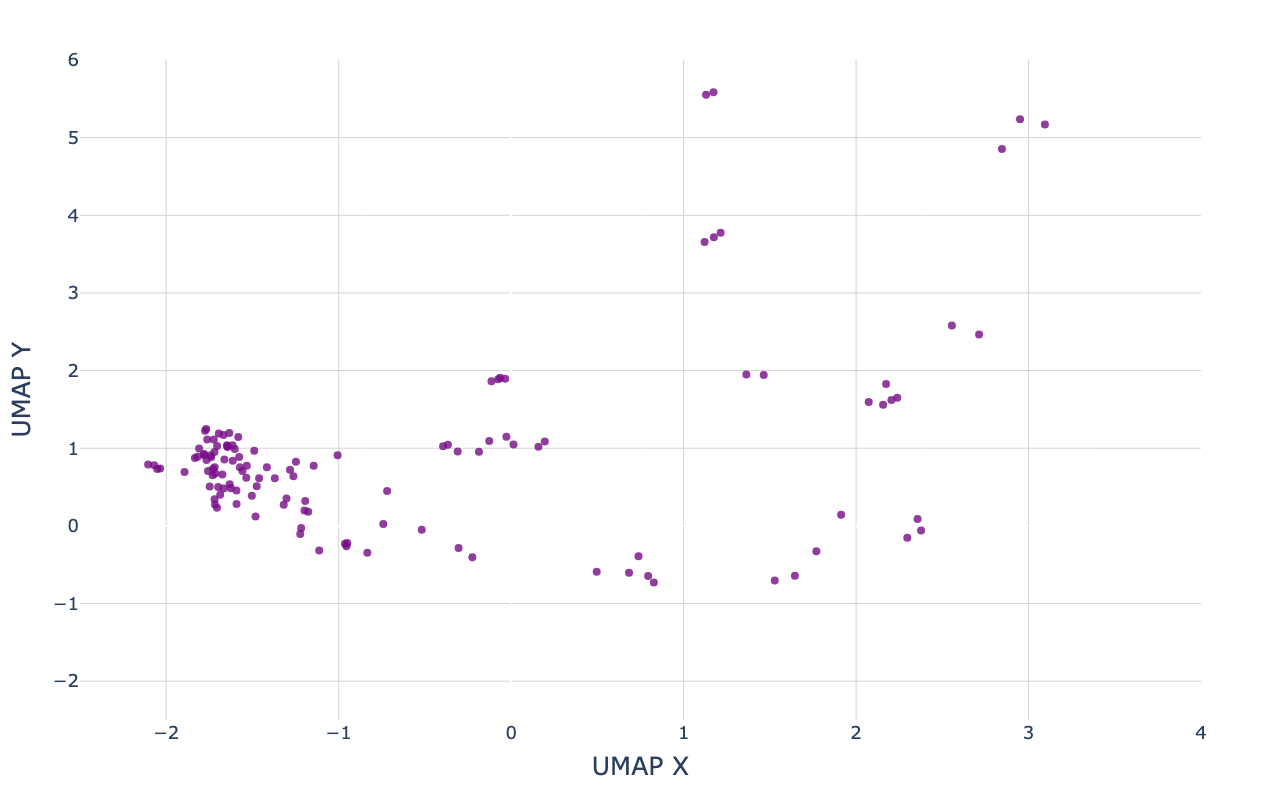}}
    \caption{Sup Heathland}
    \end{subfigure} 
    \begin{subfigure}{.48\linewidth}
    \centering
    \centerline{\includegraphics[height=4.5cm]{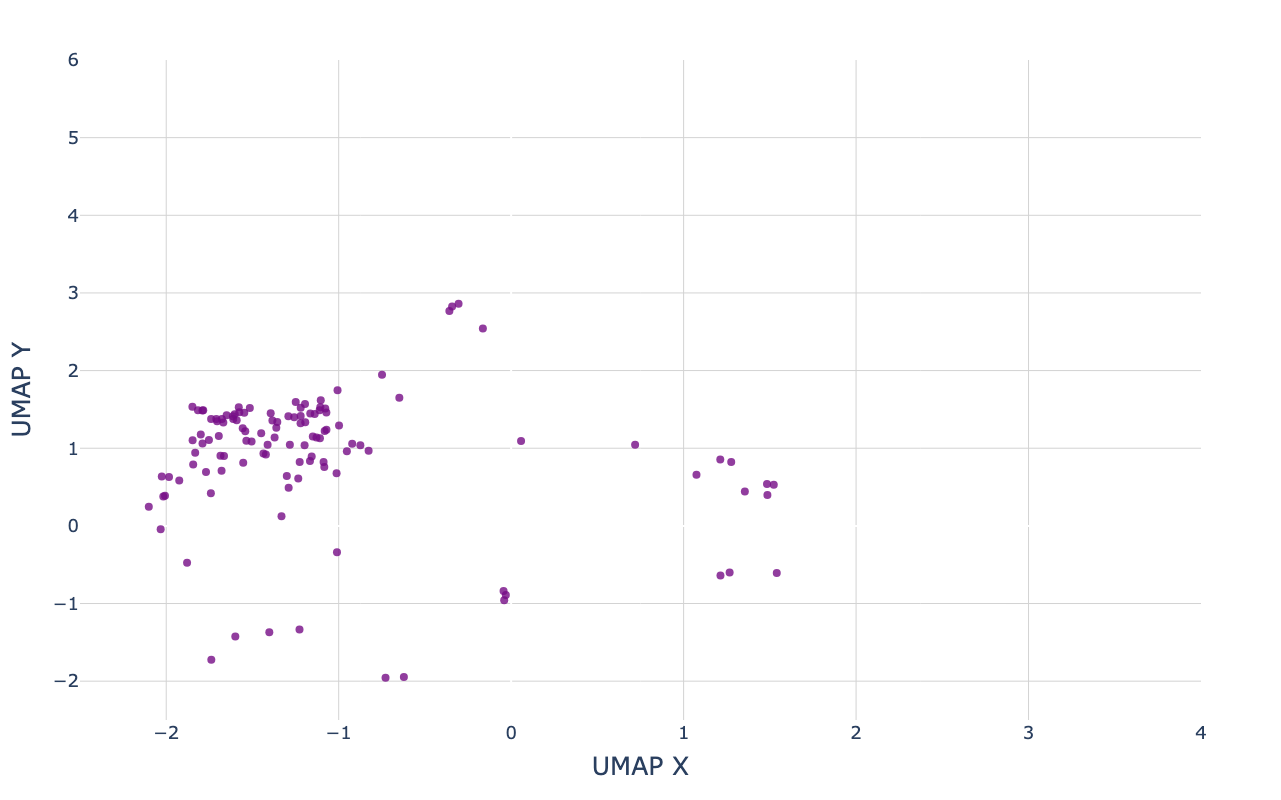}}
    \caption{SupCon Heathland}
    \end{subfigure}

    \begin{subfigure}{.48\linewidth}
    \centering
    \centerline{\includegraphics[height=4.5cm]{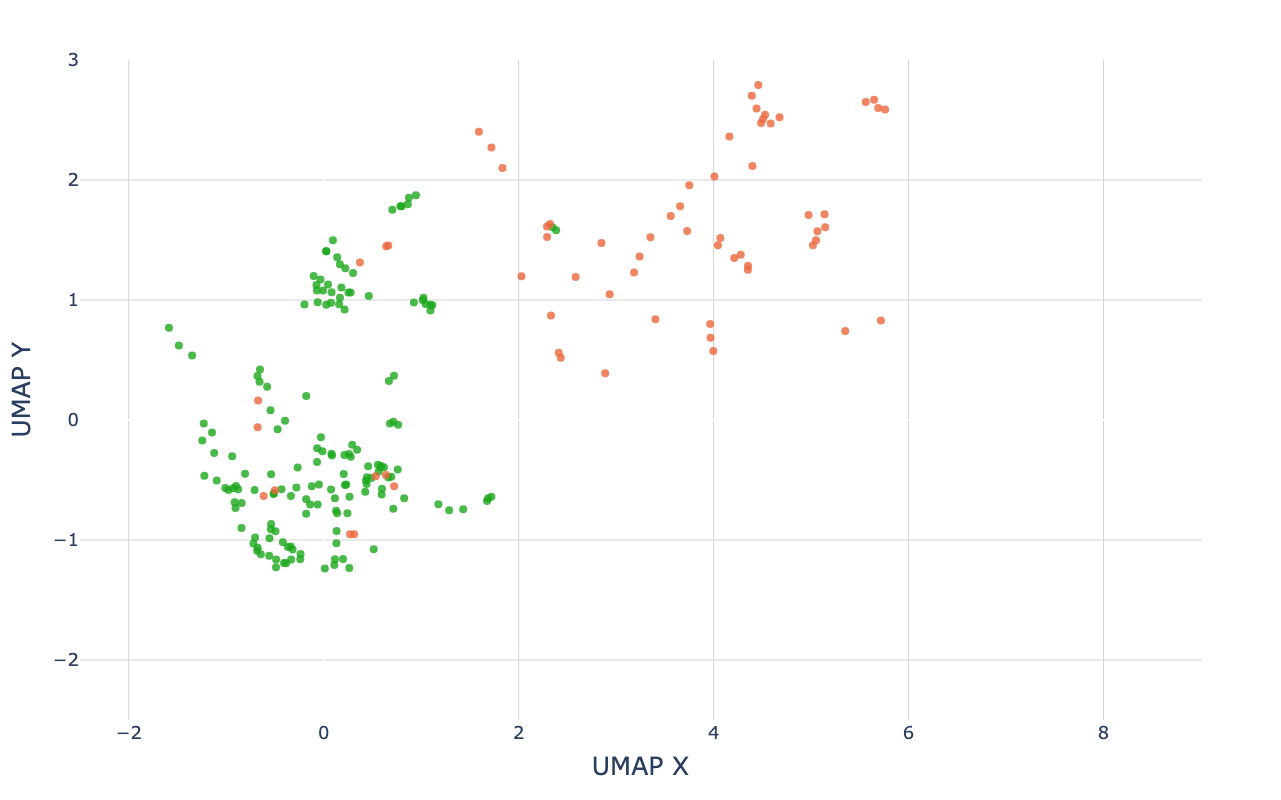}}
    \caption{Sup Wetland}
    \end{subfigure} 
    \begin{subfigure}{.48\linewidth}
    \centering
    \centerline{\includegraphics[height=4.5cm]{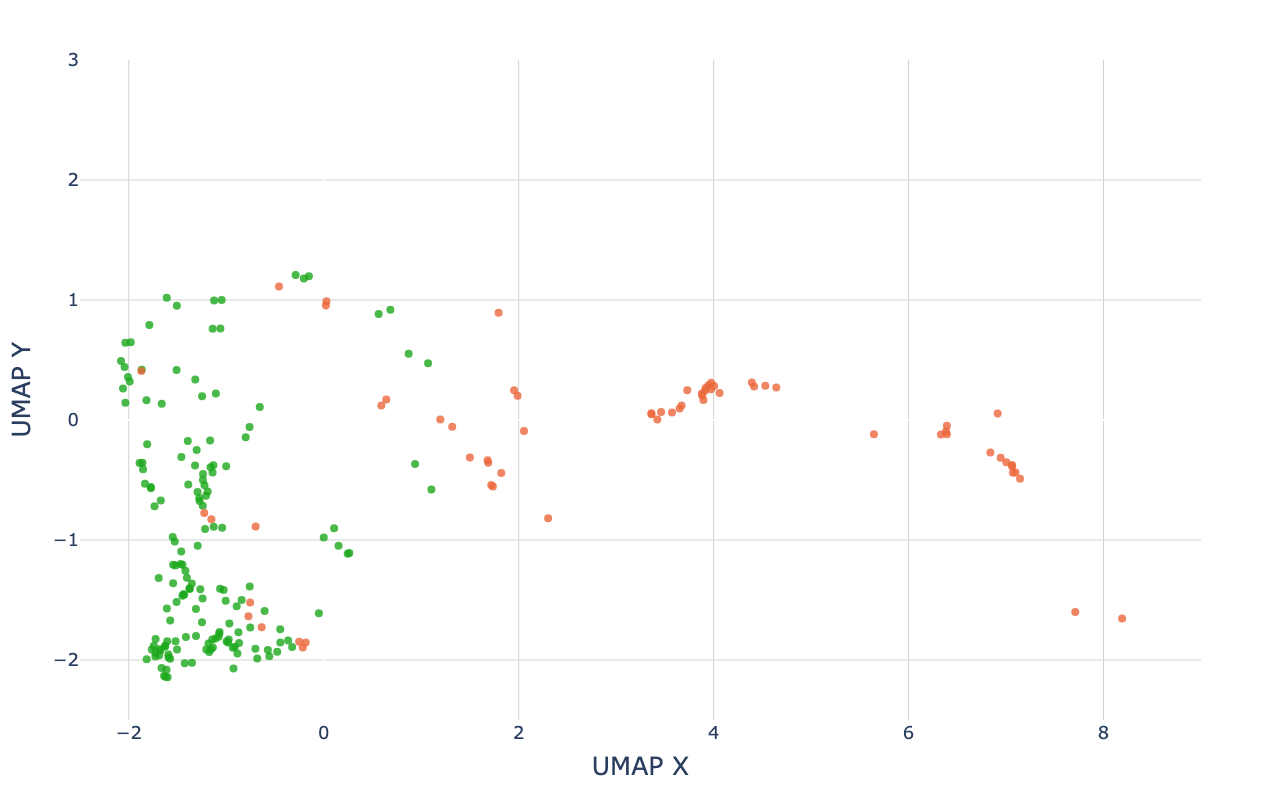}}
    \caption{SupCon Wetland}
    \end{subfigure}

  \caption{UMAPs of the embedding space produced by the image encoder on test set. SupCon is found to generate tightly clusters that are more separable from each other compared to supervised learning. Habitat symbols in this figure:
  \textcolor[HTML]{2E91E5}{$\bullet$} Acid Grassland;\textcolor[HTML]{E15F99}{$\bullet$} Arable and Horticulture;\textcolor[HTML]{1CA71C}{$\bullet$} Bog; \textcolor[HTML]{FB0D0D}{$\bullet$} Bracken; \textcolor[HTML]{DA16FF}{$\bullet$} Broadleaved Mixed and Yew Woodland; \textcolor[HTML]{222A2A}{$\bullet$} Calcareous Grassland;\textcolor[HTML]{B68100}{$\bullet$} Coniferous Woodland; \textcolor[HTML]{750D86}{$\bullet$} Dwarf Shrub Heath; \textcolor[HTML]{EB663B}{$\bullet$} Fen, Marsh, Swamp; \textcolor[HTML]{511CFB}{$\bullet$} Improved Grassland; \textcolor[HTML]{00A08B}{$\bullet$} Inland Rock; \textcolor[HTML]{FB00D1}{$\bullet$} Littoral Sediment;  \textcolor[HTML]{FC0080}{$\bullet$} Montane; \textcolor[HTML]{B2828D}{$\bullet$} Neutral Grassland; \textcolor[HTML]{6C7C32}{$\bullet$} Supra-littoral Sediment; \textcolor[HTML]{778AAE}{$\bullet$} Urban.}
  \label{fig: umaps}
\end{figure}

Figure~\ref{fig: umaps} visualise the embeddings of test images extracted by the SwinT-B trained with supervised learning and SupCon in a 2-D space with UMAP. Overall, the image encoder trained with SupCon generates tightly grouped clusters that are more separable from each other compared to supervised learning. However, SupCon appears to make the heathland cluster less tight compared with supervised learning. This reduced effect of contrastive learning could attribute to there are not enough positive samples for heathland to compare in a batched data, as heathland has relatively fewer samples in the dataset and we choose a small batch size due to computation limitations. Further, Table~\ref{tab: CHI and DBI for embedding clusters} reports the CHI and DBI of SupCon and supervised learning, indicating that SupCon reduces intra-class variances and increases inter-class variances among overall habitat clusters, grassland clusters, and woodland clusters. However, the cluster quality of wetlands is slightly penalised by SupCon. The tradeoff effect suggests SupCon method alone cannot completely address the misclassification problem brought by visually similar habitats.  

\begin{table}
    \centering
    \caption{Comparison of Calinski–Harabasz (CH) Index and Davies–Bouldin (DB) Index for supervised learning (Sup) and SupCon. A higher CHI or a lower DBI implies a better clustering quality, with low intra-class variances and high inter-class variances. As CHI and DBI must be calculated based on at least two clusters (classes), major L2 habitats with only a single L3 class, such as heathland and cropland, are excluded.}
    \begin{tabular}{lcc|cc}
        \toprule
                        & \multicolumn{2}{c}{\textbf{CH Index} \contour{black}{$\uparrow$}}  & \multicolumn{2}{c}{\textbf{DB Index} \contour{black}{$\downarrow$}} \\
        \cmidrule(r){2-3} \cmidrule(l){4-5}
        \textbf{Dataset}         & \textbf{Sup}        & \textbf{SupCon}        & \textbf{Sup}        & \textbf{SupCon} \\
        \midrule
        Overall habitats  &  407.6  &  \textbf{450.4}  &  4.50  &  \textbf{3.20} \\ \midrule
        Grasslands        &  129.2  &  \textbf{202.9}  &  3.50  &  \textbf{2.34} \\
        Woodlands        &  42.1  &  \textbf{58.5}  &  1.60  &  \textbf{1.27} \\
        Wetlands        &  \textbf{268.9}  &  212.8  &  \textbf{0.79}  &  0.80 \\
        \bottomrule
    \end{tabular}
    \label{tab: CHI and DBI for embedding clusters}
\end{table}

\FloatBarrier

\section{Effects of SupCon on CNN Backbones} \label{appendix SupCon on CNNs}
\begin{figure}[htbp]
  \begin{center}
    \includegraphics[width=0.7\linewidth]{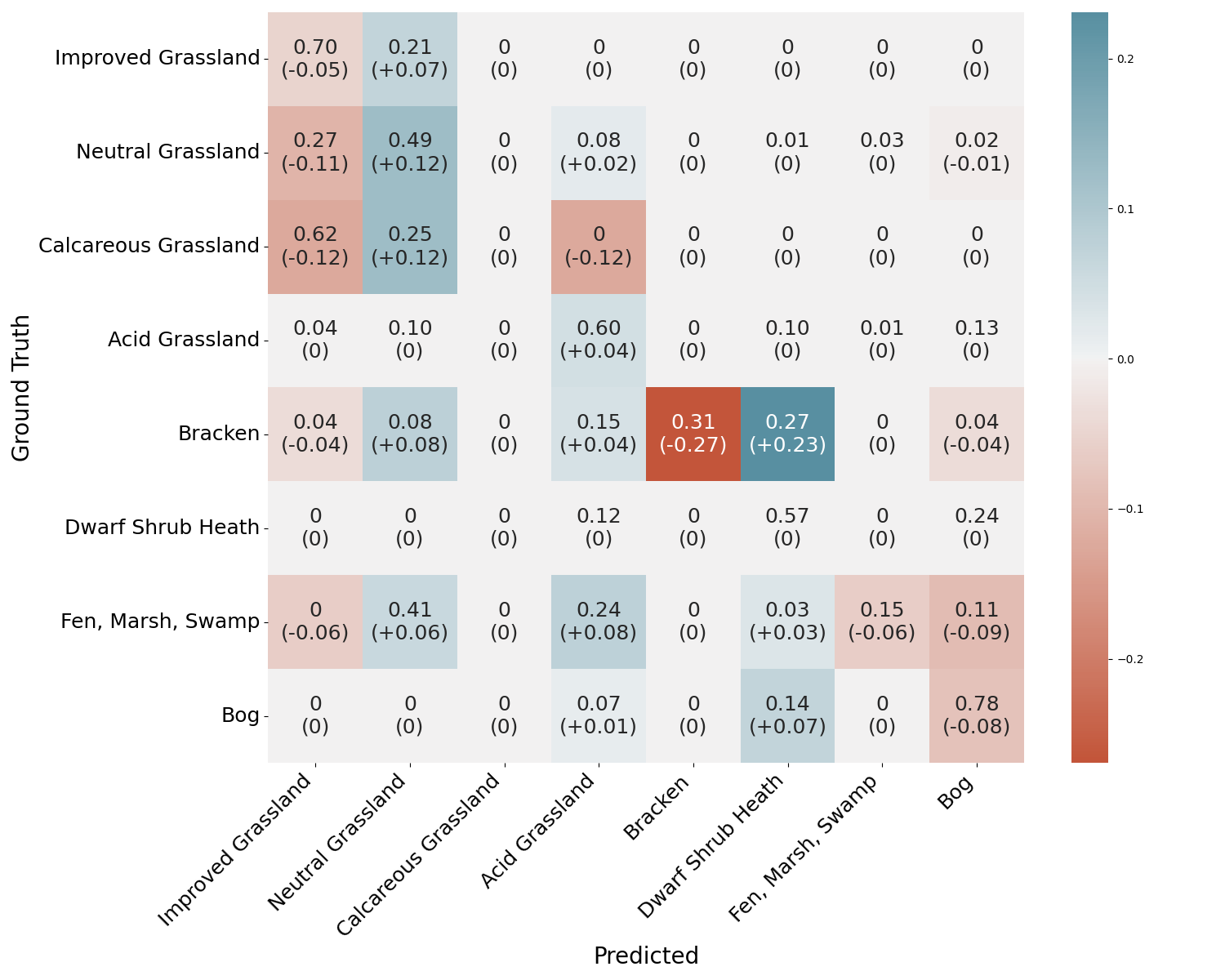}
  \end{center}
  \caption{Delta CM of WRN-50-2 with SupCon: based on the CM produced by SupCon.}
  \label{fig: delta cm of wrn-50-2}
\end{figure}

\begin{figure}[htbp]
  \begin{center}
    \includegraphics[width=0.7\linewidth]{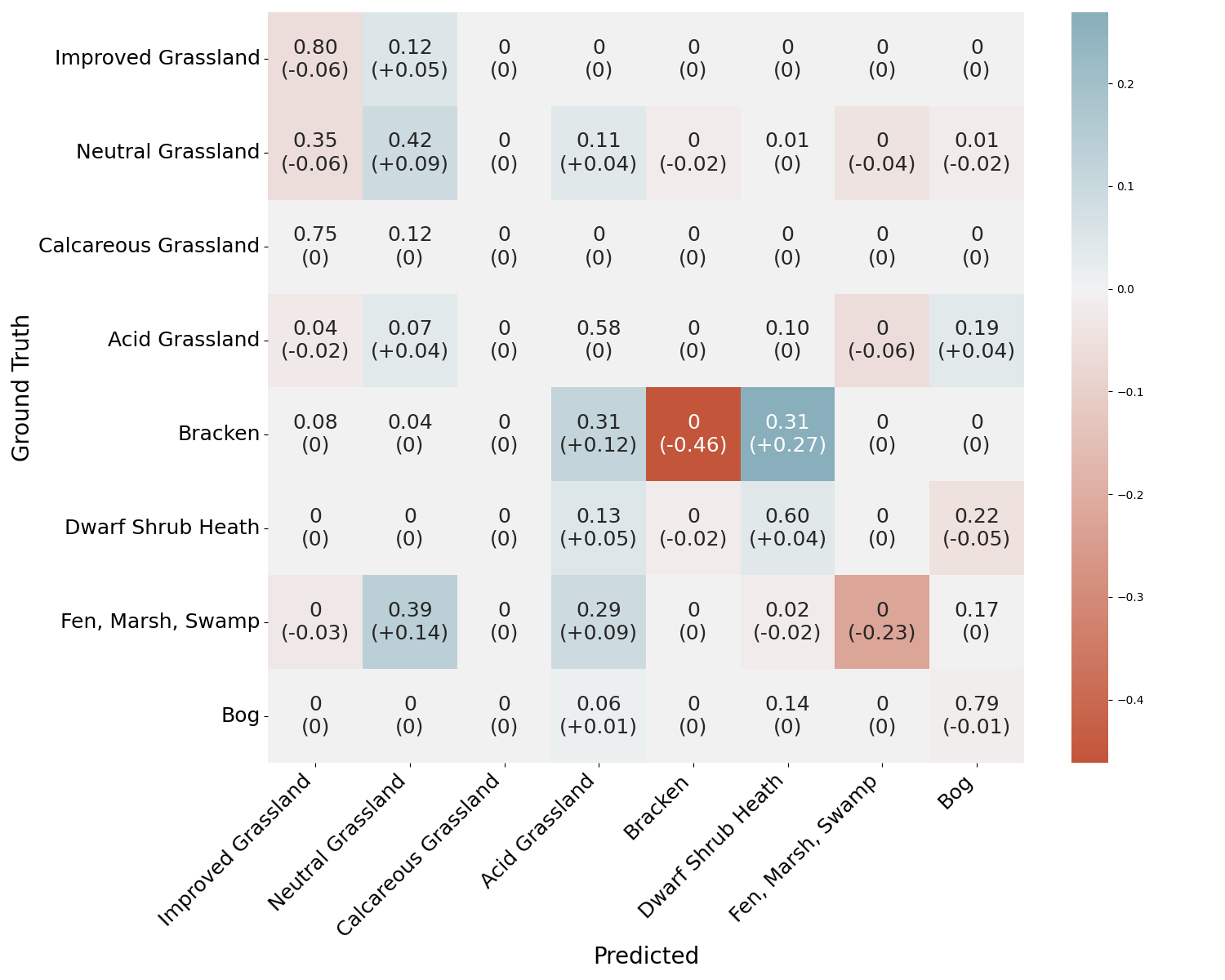}
  \end{center}
  \caption{Delta CM of ResNeXt50 with SupCon: based on the CM produced by SupCon.}
  \label{fig: delta cm of resnext50}
\end{figure}

\begin{figure}[htbp]
  \begin{center}
    \includegraphics[width=0.7\linewidth]{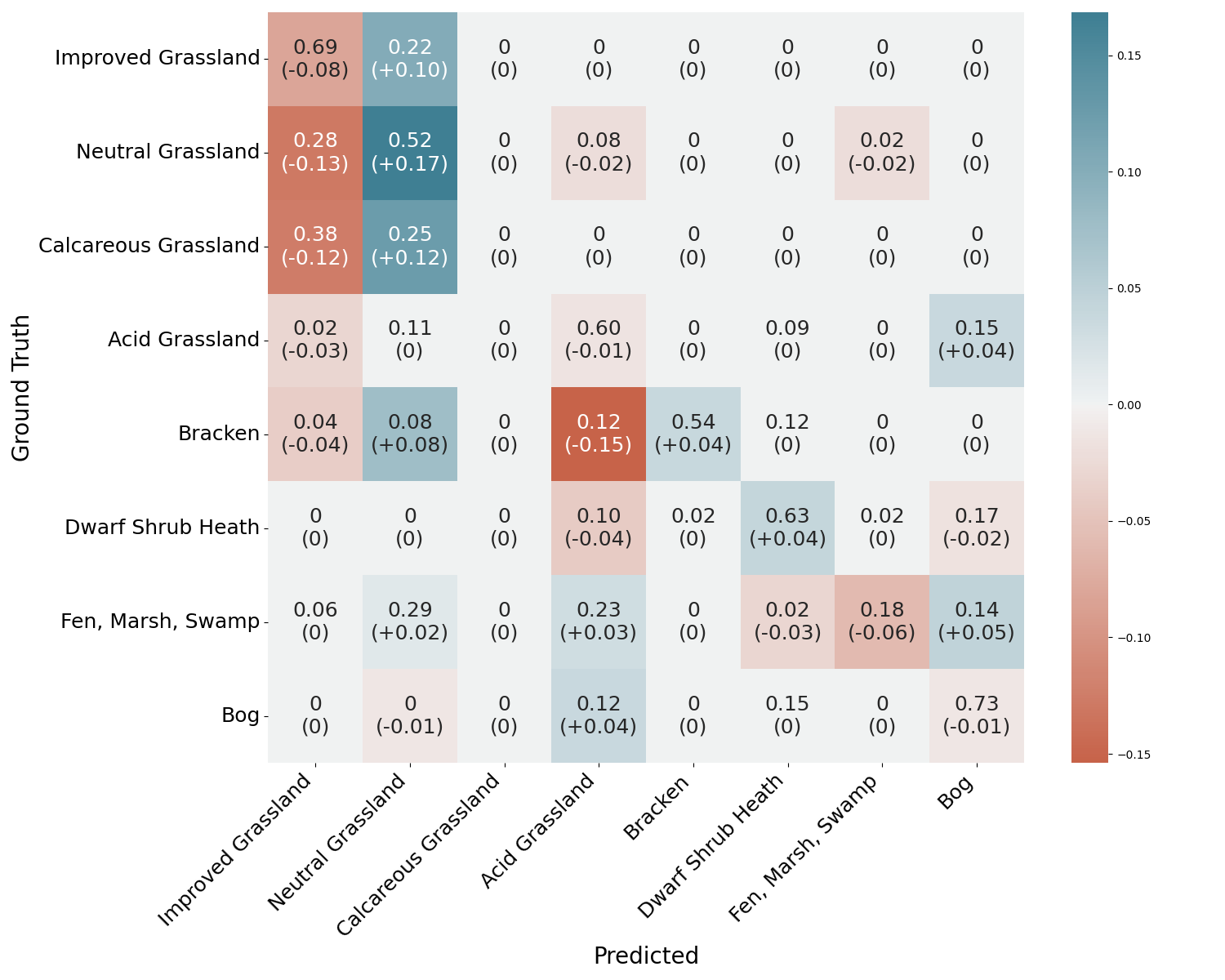}
  \end{center}
  \caption{Delta CM of EfficientNetV2-L with SupCon: based on the CM produced by SupCon.}
  \label{fig: delta cm of effl}
\end{figure}

For our employed CNNs, SupCon reduces misclassification for certain classes but increases it for others, leading to minimal or even negative overall improvement. Notably, the confusion matrices (Figures \ref{fig: delta cm of wrn-50-2}, \ref{fig: delta cm of resnext50}, \ref{fig: delta cm of effl}) show that SupCon boosts recall for Neutral Grassland but increases false positives, especially making more Improved Grassland misclassified as Neutral. In our case, SupCon tightens within-class embeddings but also pulls some samples toward incorrect clusters. 

Our underlying hypotheses for this observation are twofold. (i) Due to locality bias, CNNs may discard visual cues needed to discriminate habitats, yielding embeddings that are indistinguishable. (ii) Some anchor samples are poor class representatives for contrastive learning (e.g., images that plausibly resemble both Neutral and Improved Grassland). To address (i), we can replace CNNs with ViTs to obtain globally informed, more expressive embeddings that better suit SupCon. To address (ii), we can actively select anchors with low model uncertainty (e.g., minimum-entropy criterion) for SupCon. 

\section{Analysis of the Human-Expert Benchmark} \label{appendix human expert}

\begin{figure}[htbp]
  \begin{center}
    \includegraphics[width=0.7\linewidth]{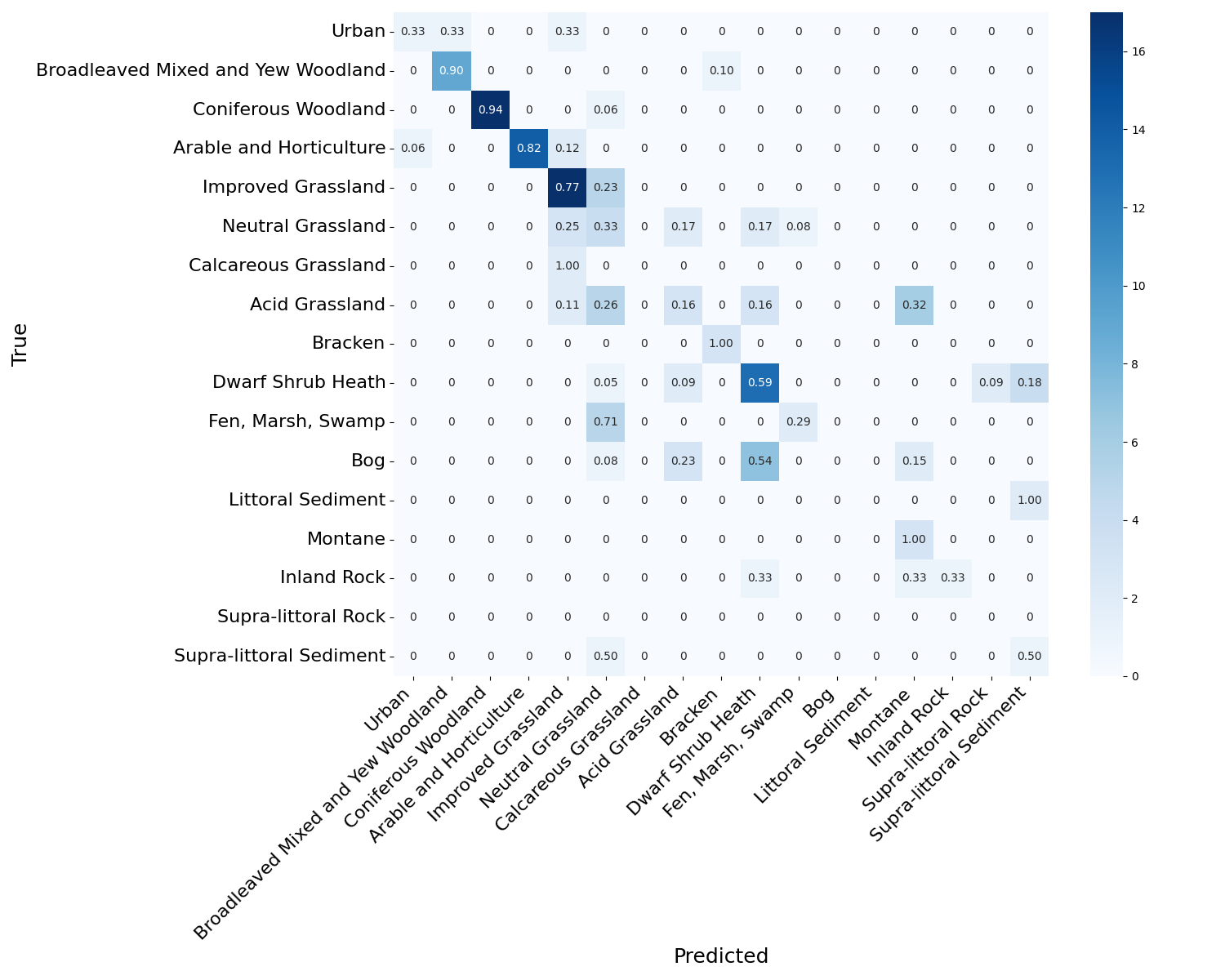}
  \end{center}
  \caption{Confusion matrices for Expert 1 performance against the ground truth, using a subset of 158 test images.}
  \label{fig: cm experts vs model--expert 1}
\end{figure}

\begin{figure}[htbp]
  \begin{center}
    \includegraphics[width=0.7\linewidth]{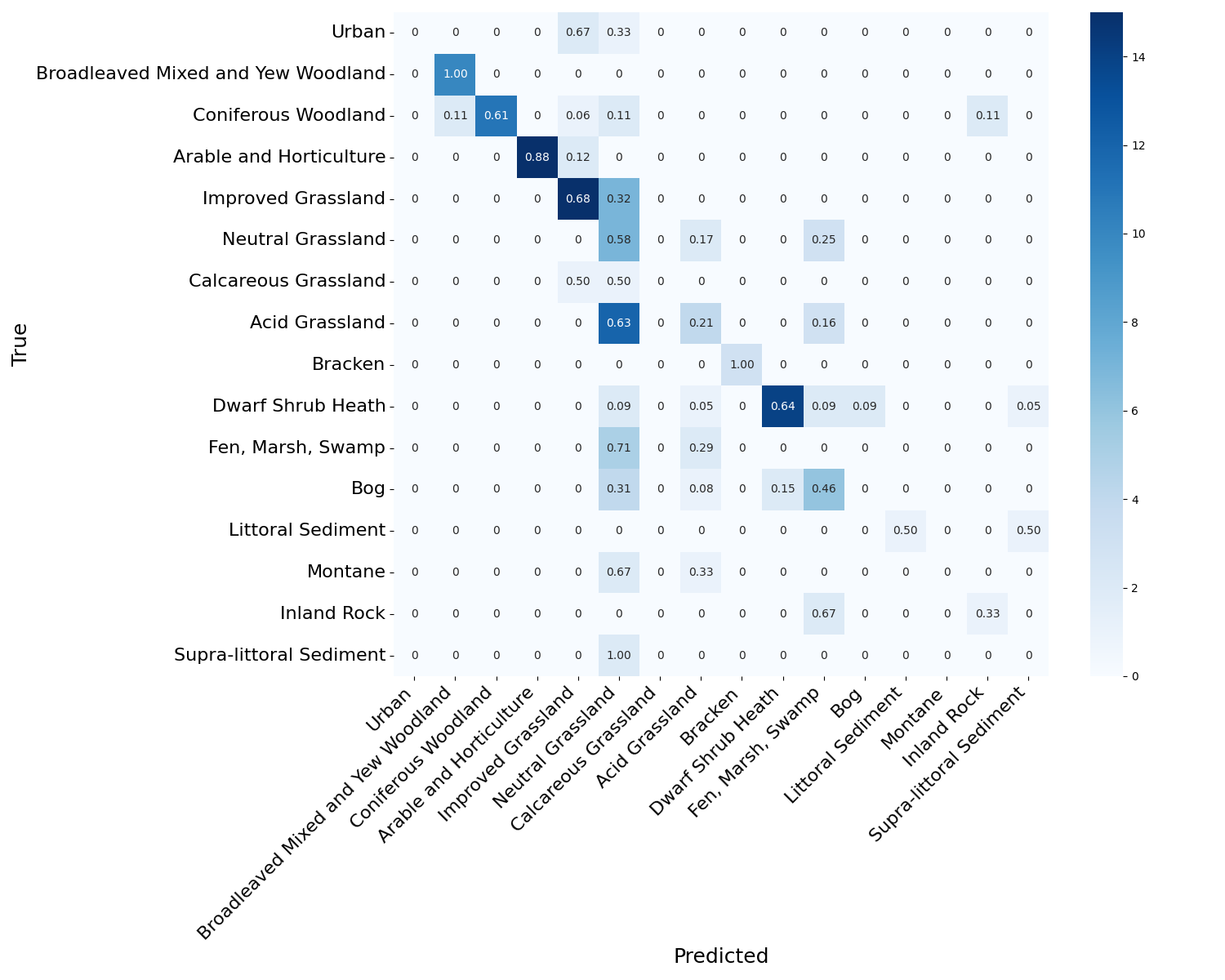}
  \end{center}
  \caption{Confusion matrices for Expert 2 performance against the ground truth, using a subset of 158 test images.}
  \label{fig: cm experts vs model--expert 2}
\end{figure}

\begin{figure}[htbp]
  \begin{center}
    \includegraphics[width=0.7\linewidth]{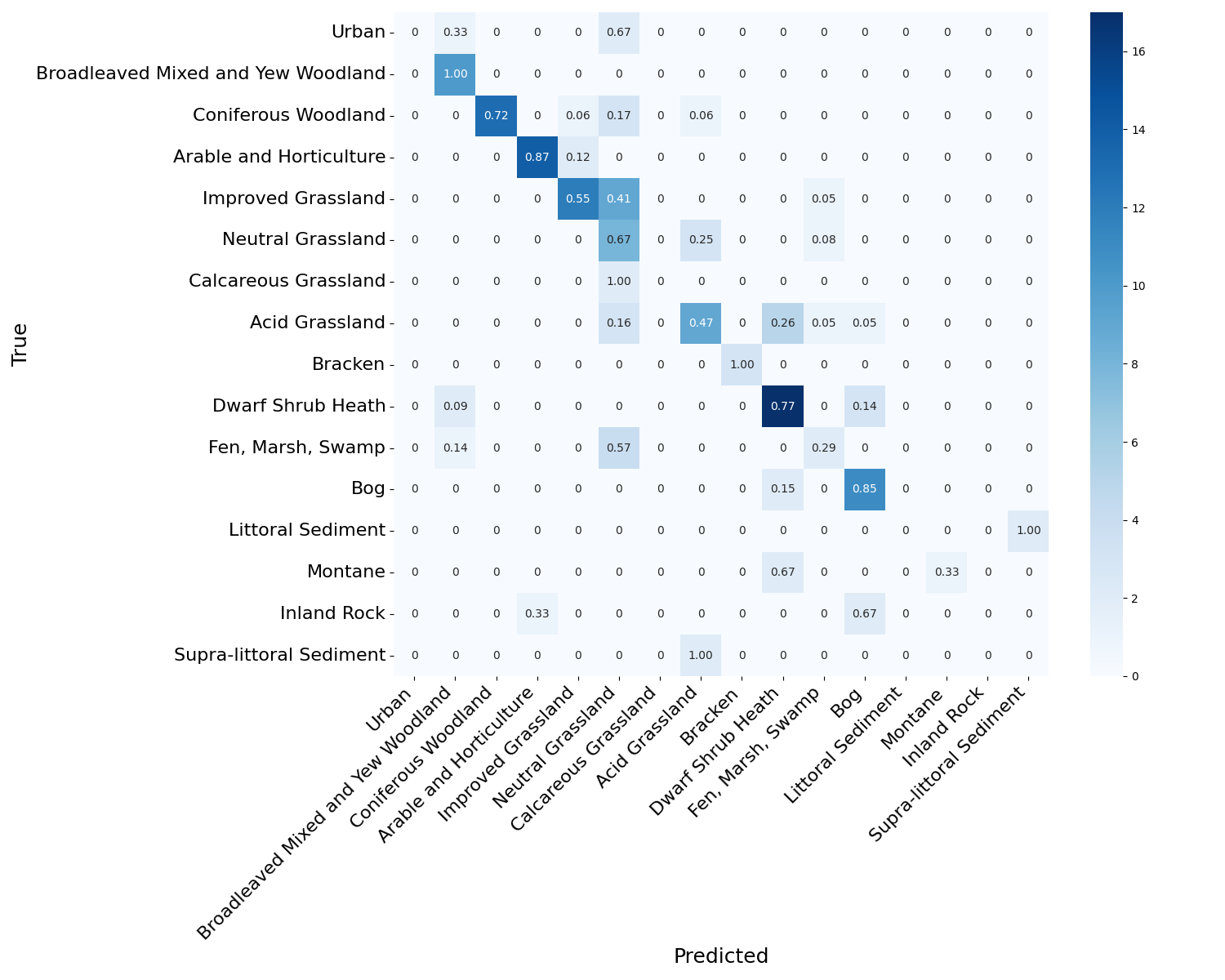}
  \end{center}
  \caption{Confusion matrices for Expert 3 performance against the ground truth, using a subset of 158 test images.}
  \label{fig: cm experts vs model--expert 3}
\end{figure}

\begin{figure}[htbp]
  \begin{center}
    \includegraphics[width=0.7\linewidth]{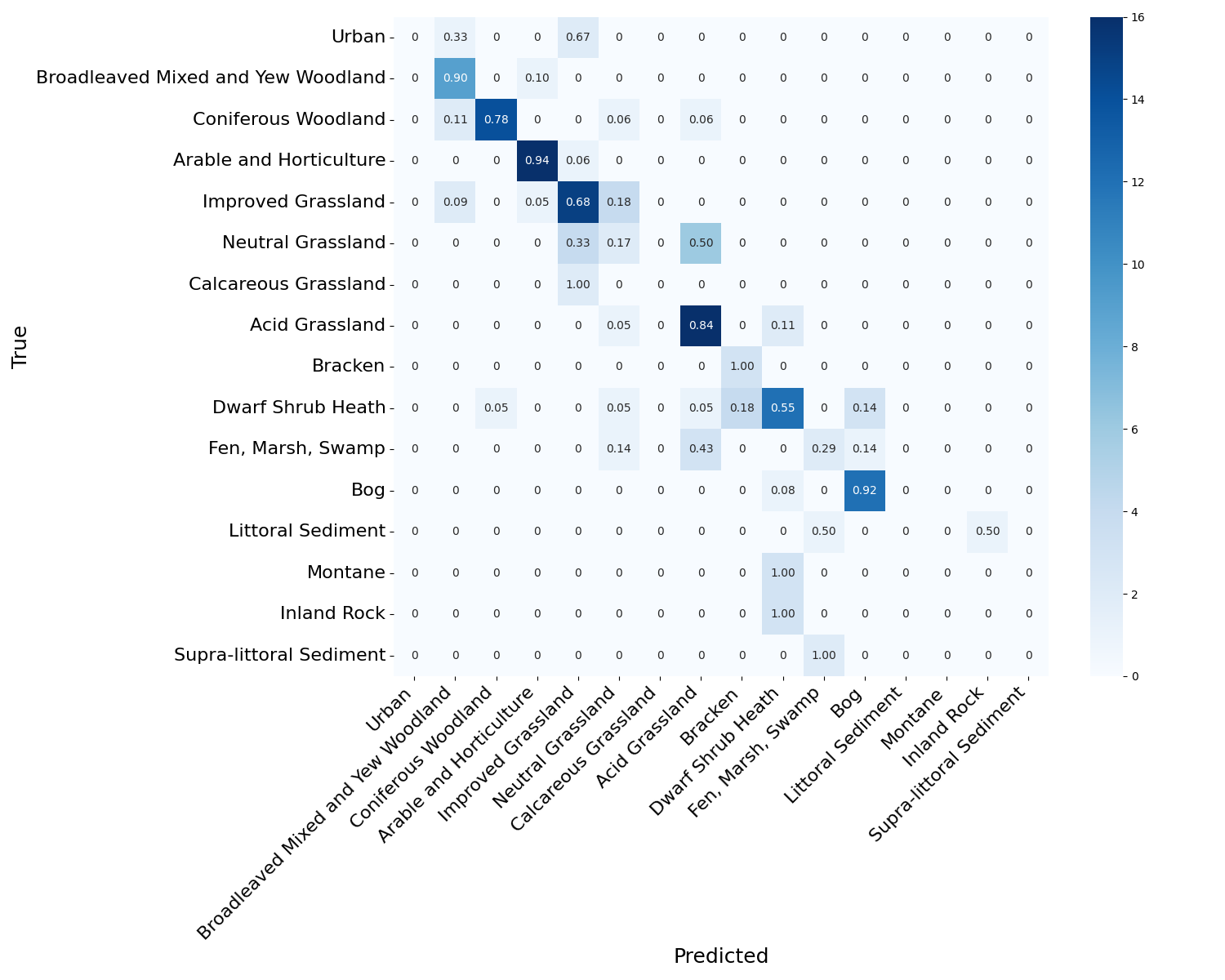}
  \end{center}
  \caption{Confusion matrices for SwinT-B (SupCon) performance against the ground truth, using a subset of 158 test images.}
  \label{fig: cm experts vs model--model}
\end{figure}

Figures \ref{fig: cm experts vs model--expert 1}, \ref{fig: cm experts vs model--expert 2}, \ref{fig: cm experts vs model--expert 3}, \ref{fig: cm experts vs model--model} present confusion matrices for three human experts and our best model (SwinT-B + SupCon) on the representative CS subset described in Section 3.4.3 of the main paper, allowing a habitat-by-habitat comparison at Level 3. Four main observations are provided:

\begin{itemize}
    \item As for the model, differentiating grassland subtypes is difficult for experts. All three experts frequently confused Improved Grassland with Neutral Grassland (up to 41\%), and a substantial fraction of Neutral Grassland was interpreted as Acid Grassland. On heathland, however, the top expert surpassed the model, correctly identifying 77\% of samples versus 55\% made the model.
    \item Interpretations diverge considerably among experts, most notably for wetland classes. Experts 1 and 2 misclassified every Bog image, whereas Expert 3 recognised 85\% correctly. By contrast, for Fen, Marsh \& Swamp (FMS) both Experts 1 and 3 achieved 29\% accuracy, while Expert 2 failed to identify any instance. The wetland performance of the model is close to that of Expert 3.
    \item Woodland and cropland receive consistently strong predictions from both humans and the model. This likely reflects their limited within-class variability (two woodland sub-classes, one cropland class) and distinctive visual features, in contrast to the five visually similar grassland sub-classes.
    \item For rarer habitats such as Littoral Sediment and Inland Rock, the experts outperform the model; the model makes no correct predictions. The shortage of training examples restricts its ability to generalise, whereas humans can draw on prior knowledge and effective few-shot reasoning.
\end{itemize}

In summary, experts face many of the same challenges as the model, particularly when visually similar classes, e.g. grassland sub-types, are involved. The pronounced disagreement among experts indicates that single ground-level photographs often provide insufficient information for consistent judgement; supplying additional contextual data (season, location, ancillary views) could improve consensus. Finally, the model’s shortfall on minor habitats underscores the need for more balanced training data: with adequate samples, it may surpass expert performance, given that it already matches them on the majority of classes in our benchmark.

\FloatBarrier

\section{Habitat Description in UKHab} \label{appendix habitat description}

\begin{center}
\begin{longtable}{@{}l|p{10cm}@{}}
  \caption{Habitat descriptions defined in UKHab \citep{UKHab2023}. Habitats in bold belong to level 2, followed by their level 3 subcategories.}
  \label{tab:L3_habitats} \\

  \toprule
  \textbf{Habitat Name (Level)} & \textbf{Description} \\
  \midrule
  \endfirsthead

  \multicolumn{2}{@{}l}{\small\slshape Table \thetable\ continued from previous page} \\  
  \toprule
  \textbf{Habitat Name (Level)} & \textbf{Description} \\
  \midrule
  \endhead

  \midrule
  \multicolumn{2}{r}{\small\slshape Continued on next page} \\
  \endfoot

  \bottomrule
  \endlastfoot

    \textbf{Grassland (L2)}  & Total vegetation cover variable from 25--100\% -- not on waterlogged soils. Vegetation is \(\geq 75\%\) herbaceous (grasses, sedges, rushes, ferns and forbs) rather than woody, with halophytic species absent or occasional. \\ \cmidrule(lr){2-2}
    Acid Grassland (L3)  & Vegetation dominated by grasses and herbs on a range of lime-deficient soils that have been derived from acidic bedrock or from superficial deposits such as sands and gravels. Such soils usually have a low base-status, with a pH of \(<5.5\). \\ \cmidrule(lr){2-2} 
    Bracken (L4)\footnote{Bracken is treated as a L3 habitat in the CS dataset as explained in Section 3.1}  & Land with Bracken Pteridium aquilinum at \(>95\%\) canopy cover at the height of the growing season. \\ \cmidrule(lr){2-2}
    Calcareous Grassland (L3)  & Vegetation dominated by grasses and herbs on shallow, well-drained soils that are rich in bases (principally calcium carbonate) formed by the weathering of chalk and other types of limestone or base-rich rock. \\ \cmidrule(lr){2-2}
    Improved Grassland (L3) \footnote{Improved Grassland is a Broad Habitat that originates from the UK BAP Priority Habitats system \citep{jncc_bap_habitats} and is used as equivalent to an L3 habitat (Modified Grassland) in UKHab.}  & This broad habitat type is characterised by vegetation dominated by a few fast-growing grasses on fertile, neutral soils. It is frequently characterised by an abundance of rye-grass Lolium spp. and white clover Trifolium repens. Improved grasslands are typically either managed as pasture or mown regularly for silage production or in non-agricultural contexts for recreation and amenity purposes. \\ \cmidrule(lr){2-2}
    Neutral Grassland (L3)  & Vegetation dominated by grasses and herbs on a range of neutral soils, usually with a pH of \(4.5\text{--}6.5\). \\ \midrule
    
    \textbf{Woodland and Forest (L2)} & Land with \(\geq 25\%\) cover of trees that are \(\geq 5\,\text{m}\) in height. \\  \cmidrule(lr){2-2}
    Broadleaved and Mixed Woodland (L3) & Vegetation dominated by trees that are \(>5\,\text{m}\) high when mature, which form a distinct, although sometimes open, canopy with a canopy cover of \(>25\%\). It includes stands of both native and non-native broadleaved tree species and Yew \textit{Taxus baccata} where the percentage cover of these trees in the stand is \(>20\%\) of the total cover of the trees present. \\ \cmidrule(lr){2-2}
    Coniferous Woodland (L3) & Vegetation dominated by trees that are \(>5\,\text{m}\) high when mature, which form a distinct, although sometimes open, canopy that has a cover of \(>25\%\). It includes stands of both native and non-native coniferous tree species (with the exception of Yew \textit{Taxus baccata}) where the percentage cover of these trees in the stand is \(>80\%\) of the total cover of the trees present. \\ \midrule
    
    \textbf{Heathland and Shrub (L2)} & Vegetation with a \(>25\%\) cover of dwarf shrub species that are \(<1.5\,\text{m}\) high or woody species \(\leq 5\,\text{m}\) high. \\ \cmidrule(lr){2-2}
    Dwarf Shrub Heath (L3) & Vegetation that has a \(>25\%\) cover of plant species from the heath family (ericoids), Dwarf Gorse Ulex minor or Western Gorse Ulex gallii. \\ \midrule
    
    \textbf{Wetland (L2)} & A vegetated habitat that is waterlogged or inundated. \\ 
    Bog (L3) & Rain-fed (ombrotrophic) inundated or waterlogged habitats where peat has formed. \\ \cmidrule(lr){2-2}
    Fen Marsh and Swamp (L3) & Characterised by a variety of vegetation types that are found on minerotrophic (groundwater-fed), permanently, seasonally or periodically waterlogged peat, peaty soils or mineral soils. Fens are peatlands which receive water and nutrients from groundwater and surface run-off, as well as from rainfall. Flushes are associated with lateral water movement, and springs with localised upwelling of water. Marsh is a general term usually used to imply waterlogged soil; it is used more specifically here to refer to fen meadows and rush-pasture communities on mineral soils and shallow peats. Swamps are characterised by tall emergent vegetation. Reedbeds (i.e. swamps dominated by stands of Common Reed Phragmites australis) are also included. \\ \midrule
    
    \textbf{Cropland (L2)} & Agricultural or horticultural land that has been cultivated or cropped within the current or previous year or left as fallow as part of an active arable rotation. \\ \cmidrule(lr){2-2}
    Arable and Horticulture & Arable cropland (including perennial woody crops and intensively managed commercial orchards), commercial horticultural land (such as nurseries, commercial vegetable plots and commercial flower-growing areas), freshly ploughed land, leys, rotational set-aside and fallow. \\ \midrule
    
    \textbf{Urban (L2 \& L3)\footnote{Urban is a L2 habitat in the UKHab with a single L3 subtype. However, the CS dataset actually treats Urban as one of the L3 classes. So here we denote Urban as a class at both L2 and L3.}} & Constructed, industrial and other artificial habitats. Urban and rural settlements, farm buildings, caravan parks and other human-made built structures such as industrial estates, retail parks, waste and derelict ground, urban parkland and urban transport infrastructure. \\ \midrule
    
    \textbf{Sparsely Vegetated Land (L2)} & Unvegetated, disturbed (regularly or drastically periodically) or sparsely vegetated habitats (permanently or periodically naturally unvegetated areas) that are inhabited by stress-tolerating vegetation with cover \(<50\%\). \\ \cmidrule(lr){2-2}
    Inland Rock (L3) & Natural and artificial exposed rock surfaces that are mappable, such as inland cliffs, caves, screes and limestone pavements, as well as various forms of excavations and waste tips, such as quarries and quarry waste. \\ \cmidrule(lr){2-2}
    Supra-littoral Rock (L3) & The region of rocky shore, including cliffs and slopes, that is immediately above the highest water level and subject to wetting by spray or wave splash (also called the ‘splash zone’). Features that may be present include vertical rock, boulders, gullies, ledges and pools, depending on the wave exposure of the site and its geology. \\ \cmidrule(lr){2-2}
    Supra-littoral Sediment (L3) & Sand and shingle coastal habitats that are above the highest water level and subject to wetting by regular or occasional spray or wave splash. \\ \midrule

    \textbf{Marine Inlets and Transitional Waters (L2)} & Intertidal habitats on various substrates and with water of variable salinity between Mean High Water Mark and Mean Low Water Mark around the UK coast. \\ 
    Littoral Rock (L3) & The geology and wave exposure of the shore influence the form, which can be as varied as vertical rock, shore platforms, boulder shores or rocky reefs surrounded by areas of sediment. These two factors are also major influences on the associated marine communities. Relatively soft rock, such as chalk and limestone, can support boring species whereas colonisation of basalt and granite is limited to the rock surfaces. In all cases, there is a distinct zonation of species down the shore, which principally reflects the degree of immersion and emersion by the tide. \\ \cmidrule(lr){2-2}
    Littoral Sediment (L3) & Areas of littoral sediment are widespread around the UK, forming features such as beaches, sand banks and intertidal mudflats. A large proportion of this habitat occurs in estuaries and inlets, where it can cover extensive areas. Significant but smaller areas of littoral sediment also occur at the head of inlets and sea lochs. Beaches, which tend to be composed of sandier material, develop in more exposed situations and are also widely distributed. Sandflats are more common in northern and western parts of the country, and finer-grained flats are more common in southern and eastern areas. Muddy sediments usually occur in sheltered areas, especially estuaries. \\ \midrule

    \textbf{Montane (L3)\footnote{Montane is a Broad Habitat that originates from the UK BAP Priority Habitats system \citep{jncc_bap_habitats} and is used as equivalent to an L3 habitat in UKHab. We refer to \cite{NaturalEngland2014Montane} for its definition.}} & Montane habitats consist of a range of near-natural vegetation which lie above the natural treeline. The vegetation within these habitats includes dwarf-shrub heaths, grass-heaths, dwarf-herb communities, willow scrub, and snowbed communities. The most abundant vegetation types are heaths dominated by heather Calluna vulgaris and billbery Vaccinium myrtillus, typically with abundant bryophytes and/or lichens; and siliceous alpine and boreal grasslands with stiff sedge Carex bigelowii and moss heaths. Rarer vegetation types include snow-bed communities with dwarf willow Salix herbacea and various bryophytes and lichens, and sub-arctic willow scrub \\


\end{longtable}
\end{center}